\documentclass[sigconf]{acmart}
\AtBeginDocument{%
  \providecommand\BibTeX{{%
    \normalfont B\kern-0.5em{\scshape i\kern-0.25em b}\kern-0.8em\TeX}}}

\copyrightyear{2022}
\acmYear{2022}
\setcopyright{acmcopyright}
\acmConference[KDD '22]{Proceedings of the 28th ACM
SIGKDD Conference on Knowledge Discovery and Data Mining}{August 14--18,
2022}{Washington, DC, USA}
\acmBooktitle{Proceedings of the 28th ACM SIGKDD Conference on Knowledge
Discovery and Data Mining (KDD '22), August 14--18, 2022, Washington, DC, USA}
\acmPrice{15.00}
\acmISBN{978-1-4503-9385-0/22/08}
\acmDOI{10.1145/3534678.3539281}

\usepackage{multirow}

\usepackage{amsthm}

\usepackage{amsfonts,amssymb}
\usepackage{amsmath}
\usepackage{bm}

\newtheorem{myProb}{\textbf{Problem}}
\usepackage[linesnumbered,ruled,vlined]{algorithm2e}
\usepackage{booktabs}
\usepackage{multirow}
\usepackage{graphicx}
\usepackage[table,xcdraw]{xcolor}
\usepackage{subfigure}
\usepackage{enumitem}
\usepackage[normalem]{ulem}
\useunder{\uline}{\ul}{}

\begin{document}

\begin{sloppypar}
\title{Spatio-Temporal Graph Few-Shot Learning with Cross-City Knowledge Transfer}

% % \author{Anonymous author(s)}

\author{Bin Lu{$^{\dagger}$}, Xiaoying Gan{$^{\dagger *}$}, Weinan Zhang{$^{\dagger}$}, Huaxiu Yao{$^\S$}, Luoyi Fu{$^{\dagger}$}, Xinbing Wang{$^{\dagger}$}}
\affiliation{
    {$^{\dagger}$}Shanghai Jiao Tong University
    \country{China}
}
\affiliation{{$^\S$}Stanford University
    \country{USA}}
\email{{robinlu1209, ganxiaoying, wnzhang, yiluofu, xwang8}@sjtu.edu.cn}
\email{huaxiu@cs.stanford.edu}

\renewcommand{\shortauthors}{Bin Lu et al.}
\thanks{*Corresponding author: Xiaoying Gan <ganxiaoying@sjtu.edu.cn>}
\begin{abstract}
Spatio-temporal graph learning is a key method for urban computing tasks, such as traffic flow, taxi demand and air quality forecasting. Due to the high cost of data collection, some developing cities have few available data, which makes it infeasible to train a well-performed model. To address this challenge, cross-city knowledge transfer has shown its promise, where the model learned from data-sufficient cities is leveraged to benefit the learning process of data-scarce cities. However, the spatio-temporal graphs among different cities show irregular structures and varied features, which limits the feasibility of existing Few-Shot Learning (\emph{FSL}) methods. Therefore, we propose a model-agnostic few-shot learning framework for spatio-temporal graph called ST-GFSL. Specifically, to enhance feature extraction by transfering cross-city knowledge, ST-GFSL proposes to generate non-shared parameters based on node-level meta knowledge. The nodes in target city transfer the knowledge via parameter matching, retrieving from similar spatio-temporal characteristics. Furthermore, we propose to reconstruct the graph structure during meta-learning. The graph reconstruction loss is defined to guide structure-aware learning, avoiding structure deviation among different datasets. We conduct comprehensive experiments on four traffic speed prediction benchmarks and the results demonstrate the effectiveness of ST-GFSL compared with state-of-the-art methods.

%take traffic speed prediction as an example and conduct extensive studies on four public datasets. The experiment results show that our ST-GFSL framework achieves the state-of-the-art results.

% To balance model parameters and training samples of different scales, ST-GFSL proposes to generate non-shared parameters by learning node-level meta knowledge. Due to the similarity of node-level meta knowledge across different cities, cross-city knowledge transfer is achieved via parameter matching of feature extractors.

% To balance model parameters and training samples of different scales, ST-GFSL proposes to generate non-shared parameters through node-level meta knowledge matching.
% In order to capture the diverse features, ST-GFSL learns the node-level meta knowledge and generate non-shared model parameters for feature extraction.
% Furthermore, we introduce ST-Meta graph and graph reconstruction loss to guide structure-aware few-shot training, reducing structure deviation among different source datasets. Our experiments take traffic speed prediction as an example and conduct extensive studies on public datasets of four cities. The experiment results show that our ST-GFSL framework achieves the state-of-the-art results.
\end{abstract}
% Due to the high cost of data acquisition, existing models cannot be directly applied to new scenarios, especially data-scarce scenarios. 
% In addition, spatio-temporal data have irregular structure and dynamic feature, which limits the feasibility of few-shot learning methods. 



%%
% Keywords. The author(s) should pick words that accurately describe
% the work being presented. Separate the keywords with commas.

\maketitle

\section{Introduction}

With the rapid development of urbanization, humans, vehicles, and devices in the city have generated a considerable spatio-temporal data, which significantly change the urban management with a bunch of urban-related machine learning applications, such as traffic flow~\cite{DBLP:conf/ijcai/WuPLJZ19,DBLP:conf/cikm/LuGJFZ20}, taxi demand~\cite{DBLP:conf/aaai/YeSDF021,10.1145/3331184.3331368} and air quality forecasting~\cite{DBLP:conf/wsdm/WangZZLY21}. However, existing machine learning algorithms require sufficient samples to learn effective models, which may not be applicable to cities without sufficient data. 
% \weinan{can we add a general statistics about the ratio of data-scarce cities in the world?}
The similarity of cities inspires us to consider the cross-city knowledge transfer, which reduces the burden of data collection and improves the efficiency of smart city construction.
%Many developed cities are at the forefront of building smart cities. 
% Recent advances of spatio-temporal graph learning lead to solving a series of urban computing problems, like traffic flow~\cite{DBLP:conf/ijcai/WuPLJZ19,DBLP:conf/cikm/LuGJFZ20}, taxi demand~\cite{DBLP:conf/aaai/YeSDF021,10.1145/3331184.3331368} and air quality forecasting~\cite{DBLP:conf/wsdm/WangZZLY21}. However, their hunger for large amount of structured and labeled data limits the real-world application for developing cities where only few data are available for training. 

% 随着城市数字化的发展，一些数字化程度比较高的城市收集了大量的人、机、物的时空数据。
% 最近的一系列promising的研究采用时空图学习实现了对交通流量、空气质量、人流的预测。
% 然而上述这些方法都需要采集大量的数据进行训练才能得到不错的表现。
% 而由于城市发展的不均匀性，一些欠发达的城市仅有很少的数据可供训练。
% 城市运行规律的相似性启发了我们通过跨城市的知识迁移实现来减少新的目标城市的数据采集，提升智慧城市的建设效率。

% Currently, much research progress has been made in Few-Shot Learning (\emph{FSL}), like transfer learning~\cite{ying2018transfer}, meta learning~\cite{vanschoren2018meta}, multi-task learning~\cite{zhang2021survey}, etc. 

% in Few-Shot Learning (\emph{FSL})~\cite{ying2018transfer,vanschoren2018meta,zhang2021survey}

Currently, much research progress has been made for solving urban computing tasks in few-shot scenarios.
Wang et al. ~\cite{DBLP:conf/ijcai/WangGMLY19} model the cities as \emph{grids} and first propose to facilitate spatio-temporal prediction in data-scarce cities. In order to achieve better similar region-to-region matching, they introduce \emph{large-scale auxiliary data} (social media check-ins data). Whereas, finding and collecting the appropriate auxiliary data is inherently costly, and face the potential of risk leakage. 
In ~\cite{10.1145/2939672.2939830}, the authors propose \emph{FLORAL} to classify air quality by transfering the semantically-related dictionary from one data-rich city. 
However, knowledge transfer from one single source city faces the risk of negative transfer due to the great difference between two cities.
Yao et al.~\cite{10.1145/3308558.3313577} combine meta-learning method to learn a good initialization model from multiple source cities in target domain, however without considering the varied feature differences across cities and within cities.
More importantly, above methods are only applicable to grid-based data, but not compatible with graph-based modeling. 
Actually, the graph-based model has aroused extensive attention recently and achieved great success in spatio-temporal learning of road-network, metro-network, sensor-network data, etc.

% Wang et al. ~\cite{DBLP:conf/ijcai/WangGMLY19} first propose to facilitate spatio-temporal prediction in data-scarce cities by transferring knowledge from one data-rich city. They introduce \emph{large-scale auxiliary data} (social media check-ins data) for better similar region-to-region matching. Yao et al. ~\cite{10.1145/3308558.3313577} model the cities as \emph{grids} and combine meta-learning method to learn a good initialization model for adaptation in target domain.
% Wei et al.~\cite{10.1145/2939672.2939830} propose \emph{FLORAL} to classify air quality by transfering the semantically-related dictionary.

In this paper, our goal is to transfer the cross-city knowledge in graph-based few-shot learning scenarios, simultaneously exploring the impact of knowledge transfer across multiple cities. However, there exists following two challenges: 
\textbf{(i)} \emph{How to adapt feature extraction in target city via the knowledge from multiple source cities?} Current meta-learning methods assume the transferable knowledge to be globally shared across multiple cities. However, even in different areas of one city, the spatio-temporal characteristic varies greatly. Existing methods fail to handle the knowledge transfer among complicated scenarios effectively.
\textbf{(ii)} \emph{How to alleviate the impacts of varied graph structure on transferring among different cities?} Compared with grid-based data, graph-based modeling shows irregular structure information among cities. The edges between nodes explicitly depict various feature interactions. Existing FSL methods ignore the importance of structure when knowledge transferring, which causes unstable results and may even increase the risk of structure deviation.

To address the aforementioned challenges, we propose a novel and model-agnostic \textbf{S}patio-\textbf{T}emporal \textbf{G}raph \textbf{F}ew-\textbf{S}hot \textbf{L}earning framework called \textbf{ST-GFSL}. 
To the best of our knowledge, our work is the first to investigate the few-shot scenario in spatio-temporal graph learning. 
In order to adapt to the diversity of multiple cities, ST-GFSL no longer learns a globally shared model as usual. We propose to generate non-shared model parameters based on node-level meta knowledge to enhance specific feature extraction. 
The novel-level knowledge transfer is realized through parameter matching, retrieving from nodes with similar spatio-temporal characteristics across source cities and target city. In addition, we propose the ST-Meta Learner to learn node-level meta knowledge from both local graph structure and time series correlations. 
During the process of meta-learning, ST-GFSL proposes to reconstruct the 
graph structures of different cities based on meta knowledge.
Graph reconstruction loss is defined to guide structure-aware learning, so as to avoid the structure deviation across multiple source cities.

% ------------------------
% To transfer the knowledge to the target data-scarce city, ST-GFSL learns a well-generalized model from multiple source cities with sufficient training samples. 
% In order to adapt to different data scales and diverse characteristics among multiple cities, 
% Meanwhile, we propose to reconstruct a \emph{ST-Meta graph} and calculate \emph{graph reconstruction loss} to measure the structural deviation in cross-city knowledge transfer for structure-awareness learning. 
% ------------------------

% The training process of ST-GFSL follows MAML-based episode learning, which mimics the few-shot scenario in testing.

% to eliminates the impact of structure deviation caused by simultaneously learning on multiple graph datasets, we propose to reconstruct an \emph{ST-Meta graph} and \emph{graph reconstruction loss} for structure-awareness learning. 

% ST-GFSL generates the \emph{non-shared} model parameters to enhance feature extraction among different city datasets. Meanwhile, we introduce \emph{ST-Meta graph} and \emph{graph reconstruction loss}, which brings the benefits of structure-awareness and eliminates the impact of structure deviation on various source datasets. It is worth noting that ST-GFSL is a model-agnostic framework for tackling spatio-temporal graph few-shot learning, which can be easily combined with recent breakthroughs of spatio-temporal graph learning models. 

In summary, the main contributions of our work are as follows:
\begin{itemize}[left=1em]
    \item We investigate a challenging but practical problem of spatio-temporal graph few-shot learning in data-scarce cities. To our knowledge, we are the first to explore the few-shot scenario in spatio-temporal graph learning tasks.
    \item To address this problem, we propose a model-agnostic learning framework called ST-GFSL. ST-GFSL generates non-shared parameters by learning node-level meta knowledge to enhance the feature extraction. The novel-level knowledge transfer is achieved via parameter matching from similar spatio-temporal meta knowledge. 
    \item We propose to reconstruct the graph structure of different cities based on meta-knowledge. The graph reconstruction loss is further combined to guide structure-aware few-shot learning, which avoids the structure deviation among multiple cities.
\end{itemize}

We demonstrate the superiority of our proposed framework by the application of traffic speed prediction on four public urban datasets. Extensive experiments on METR-LA, PEMS-BAY, Didi-Chengdu and Didi-Shenzhen datasets validate the effectiveness and versatility of our approach over state-of-the-arts baselines.

\section{Related Work}
In this section, we briefly introduce the relevant research lines to our work.
% , namely spatio-temporal graph learning, graph few-shot learning and knowledge transfer across cities.

\subsection{Spatio-Temporal Graph Learning}
Spatio-Temporal Graph Learning is a fundamental and widely studied problem in urban computing tasks. In the early, researchers studied this problem from the perspective of time series analysis and put forward a series of methods such as ARIMA, VAR and Kalman Filtering~\cite{moreira2013predicting}. With the rise of deep learning and graph neural networks, graph, as an effective data structure to describe spatial structure relations, is applied to analyze a series of urban problems.
Bai et al.~\cite{DBLP:conf/ijcai/BaiYK0S19} propose STG2Seq to model multi-step citywide passenger demand based on a graph. Lu et al.~\cite{DBLP:conf/cikm/LuGJFZ20} propose spatial neighbors and semantic neighbors of road segments to capture the dynamic features of urban traffic flow. Do et al.~\cite{do2020graph} employ IoT devices on vehicles to sense city air quality and estimated unknown air pollutants by variational graph autoencoders. In particular, Pan et al.~\cite{DBLP:conf/kdd/PanLW00Z19, pan_tkde_2020} propose to leverage deep meta learning to improve the traffic prediction performance by generalizing the learning ability of different regions. 

However, the researches in the above papers are all based on a city with large-scale training data. The data-scarce scenario is not within the scope of the research, but it is an issue that is well worth investigating. In this paper, we aim to achieve spatio-temporal graph few-shot learning through cross-city knowledge transfer. In addition, we are committed to come up with a model-agnostic architecture that can be combined with latest spatio-temporal graph learning models to further improve the performance.
% As far as we know, ST-GFSL is the first to incorporate meta learning into cross-city spatio-temporal graph learning. 

\subsection{Graph Few-Shot Learning}

Few-Shot learning has yielded significant progress in the field of computer vision and natural language processing. 
% Several models and algorithms have been recently proposed and achieved positive results
, e.g., MAML~\cite{DBLP:conf/icml/FinnAL17}, Prototypical Networks~\cite{DBLP:conf/nips/SnellSZ17}, and Meta-Transfer Learning~\cite{DBLP:conf/cvpr/SunLCS19}. However, few-shot learning on non-Euclidean domains, like graph few-shot learning, has not been fully explored. Among recent few-shot learning on graphs, Meta-GNN~\cite{DBLP:conf/cikm/0002CZTZG19} is the first to incorporate meta-learning paradigm, MAML, into node classification in graphs. Nevertheless, it does not fully describe the interrelation between nodes. Liu et al.~\cite{liu2021relative} propose to assign the relative location and absolute location of nodes on graph to further capture the dependencies between nodes. Yao et al. \cite{DBLP:conf/aaai/YaoZWJWHCL20} and Ding et al. \cite{DBLP:conf/cikm/DingWLSLL20} adopt the idea of prototypical network and conduct few-shot node classification by finding the nearest class prototypes.

Through the analysis of above work, it can be found that existing methods mainly focus on few-shot node classification, while many urban computing problems are regression problems. Furthermore, compared with general attribute network, spatio-temporal graph has more complex and dynamic node characteristics. Directly combining few-shot learning methods with vanilla GNN model is infeasible to capture the complicated node correlations.

% Through the analysis of the above work, it can be found that they all focus on the classification of small sample nodes, while many urban computing problems are regression problems. Furthermore, compared with the general attribute network, the spatio-temporal graph has more complex and dynamic node characteristics, and the sampling general graph neural network cannot capture the correlation between nodes.

% However, few-shot learning on non-Euclidean domains, like graph few-shot learning, is under-investigated. 
% Among recent few-shot learning on graphs, Meta-GNN~\cite{DBLP:conf/cikm/0002CZTZG19} is the first to incorporate meta-learning paradigm, MAML, into node classification task of graph neural network, but it lacks consideration of graph characteristic. Yao et al. \cite{DBLP:conf/aaai/YaoZWJWHCL20}, Huang et al.\cite{DBLP:conf/nips/HuangZ20} and Ding et al. \cite{DBLP:conf/cikm/DingWLSLL20} adopted the idea of prototype network on graph node classification tasks. However, most urban computing problems are node regression problem, the method of metric-based few-shot learning framework is not suitable.
% Bose et al. \cite{bose2020metagraph} adapts gradient-based meta learning to optimize a shared parameter initialization for link prediction, but the parameter-shared model are not effective for capturing dynamic spatio-temporal features.

\subsection{Knowledge Transfer Across Cities}

Knowledge transfer addresses machine learning problems in data-scarce scenarios. Especially in urban computing tasks, how to realize knowledge transfer across cities to reduce the cost of data collection and improve learning efficiency is an ongoing research problem. \emph{FLORAL}~\cite{10.1145/2939672.2939830} is an early work that implements air quality classification by transfering knowledge from a city existing sufficient multimodal data and labels.
\emph{RegionTrans} ~\cite{DBLP:conf/ijcai/WangGMLY19} studies knowledge transfer cross-cities by dividing cities into different grids for spatio-temporal feature matching. Yao et al.~\cite{10.1145/3308558.3313577} first propose \emph{MetaST} to transfer knowledge from multiple cities. 

Neverthelss, the above research can not be directly applied to our task, mainly for the following reasons: (1) Many urban computing problems are regression problems, while \emph{FLORAL} is designed for a classification problem. (2) \emph{RegionTrans} and \emph{MetaST} are designed for grid-based data, which is incompatible for graph-based modeling in our tasks. Meanwhile, during matching process, \emph{RegionTrans} introduces additional social media check-ins data, which reduced its versatility. 
(3) \emph{FLORAL} and \emph{RegionTrans} only focus on knowledge transfer from a single source city. How to make use of the data from multiple cities and avoid negative transfer is a problem worthy of study.
In this paper, we aim to learn the cross-city meta knowledge from multiple graph-based datasets and transfer to a data-scarce city without introducing auxiliary datasets.

\begin{figure*}
    \centering
    \includegraphics[width=0.92\linewidth]{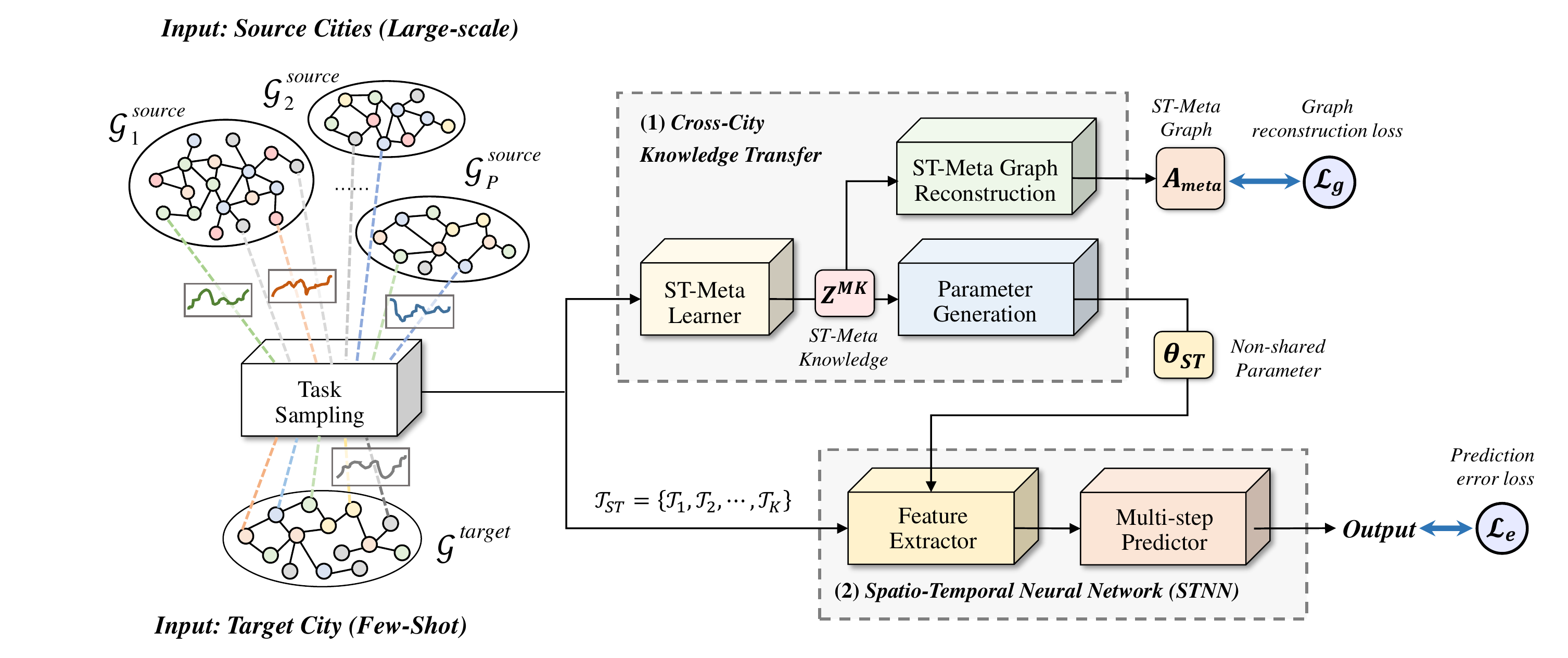}
    \caption{The framework of proposed ST-GFSL. The left side is the input of the model, in which the source cities have large-scale training samples and the target city's is few-shot. The right side shows two main parts of ST-GFSL: (1) Cross-City Knowledge Transfer and (2) Spatio-Temporal Neural Network (STNN).}
    \label{fig:framework}
\end{figure*}

\section{Preliminary}

% \subsection{Problem Formulation}
We denote the spatio-temporal graph as $\mathcal{G}_{ST}=(\mathcal{V}, \mathcal{E}, \mathbf{A}, \mathbf{X})$. (1) $\mathcal{V}=\{v_1, v_2, \cdots, v_N\}$ denotes the nodes set, and $N=\left|\mathcal{V}\right|$ is the number of nodes. (2) $\mathcal{E}=\{e_{ij}=(v_i, v_j)\} \subseteq (\mathcal{V} \times \mathcal{V})$ denotes the edges set. (3) $\mathbf{A}=\{a_{ij}\}\in \mathbb{R}^{N \times N}$ is the adjacency matrix of spatio-temporal graph. $a_{ij}=1$ indicates that there is an edge between node $v_i$ and $v_j$; otherwise, $a_{ij}=0$. 
(4) $\mathbf{X}$ is the node feature matrix, refering to the message passing on the graph, such as traffic speed, concentration of air pollutants and passenger flow of taxis over a period of time. We take the node feature observed at time $t$ as a graph signal $\mathbf{X}^{t} \in \mathbb{R}^{N \times d}$, where $d$ is the dimension of node feature. 

% (4) $\mathbf{X}\in \mathbb{R}^{N \times d}$ is node feature matrix with $x_i \in \mathbb{R}^d$ representing the spatio-temporal characteristics of a given node $v_i$.

\begin{myProb}
\textbf{Spatio-Temporal Graph Forecasting}\,
Suppose we have $T$ historical spatio-temporal graph signals, and we want to predict future $M$ graph signals. The forecasting task is formulated as learning a function $f(\cdot)$ given a spatio-temporal graph $\mathcal{G}_{ST}$:
\begin{equation}
    [\mathbf{X}^{t-T+1}, \cdot\cdot\cdot, \mathbf{X}^{t}; \mathcal{G}_{ST}] \xrightarrow{f(\cdot)} [\mathbf{X}^{t+1}, \cdot\cdot\cdot, \mathbf{X}^{t+M}]\text{.}
\end{equation}
\end{myProb}

% $\mathbf{X} = [\mathbf{X^{V_m}}, \mathbf{X^{V_a}}]$ denotes the node features, where node message feature $\mathbf{X^{V_m}}=[x_1^{V_m}, x_2^{V_m}, \cdots, x_N^{V_m}]\in \mathbb{R}^{N \times d_m}$ and node attribute feature $\mathbf{X^{V_a}}=[x_1^{V_a}, x_2^{V_a}, \cdots, x_N^{V_a}]\in \mathbb{R}^{N \times d_a}$. Node message feature refers to the messgae passing on the graph, such as traffic speed, concentration of air pollutants and passenger flow of taxis over a period of time. Node attribute feature refers to the attribute information of the node itself, such as the geographical location, POI information, etc.

\begin{myProb}
\textbf{Spatio-Temporal Graph Few-Shot Learning}\,
Suppose we have a set of $P$ source spatio-temporal graphs of data-rich cities  $\mathcal{G}^{source}_{1:P}=\{\mathcal{G}_1^{source}, \cdots, \mathcal{G}_P^{source}\}$ and a target spatio-temporal graph of data-scarce city $\mathcal{G}^{target}$. After training on $\mathcal{G}^{source}_{1:P}$, the model is capable of leveraging the meta knowledge from multiple source graphs and is tasked to predict on a disjoint target scenario, where only few-shot structured data of $\mathcal{G}^{target}$ is available.
\end{myProb}

% \subsection{spatio-temporal Graph Neural Network}

% \subsection{Graph Neural Network}
% We treat graph convolutions as a message-passing process in which information can be passed from one node to another along edges directly. Formally, the message-passing GNN is defined as follow:
% \begin{equation}
%     h_{v}^{l} = f(h_{v}^{l-1}, \sum_{u\inN(v)}g(h_{u}^{l-1}, x_{vu}^e))
% \end{equation}
% where $h_v^l$ is the $l$-th layer representation of node $v$, $N(v)$ contains the neighborhoods of node $v$, and $x_{vu}^e$ is the edge feature vector of node $(v,u)$. $f(\cdot)$ and $g(\cdot)$ are functions with learnable parameters. We initialize $h^0=X$ and the final representation $h^L$ will be used for downstream tasks.

\section{Methodology}

In this section, we describe the proposed ST-GFSL framework in detail. We first give an overview of ST-GFSL, as shown in Figure \ref{fig:framework}. The left side of the figure shows the input of ST-GFSL. We transfer the knowledge from multiple cities, and the target city only has few-shot training samples. The right side of the figure is mainly composed of two parts: \emph{Spatio-Temporal Neural Network (STNN)} and \emph{Cross-city Knowledge Transfer}. Specifically, STNN served as the base feature extractor in ST-GFSL, where any spatial-temporal learning architecture can be used, such as Graph Neural Networks (GNNs), Recurrent Neural Networks (RNNs) and other state-of-the-art spatio-temporal graph learning models. Second, \emph{Cross-City Knowledge Transfer} module transfers knowledge from multiple source cities, which are depicted in the grey dotted box of Figure \ref{fig:framework}. Concretely, we first design the \emph{ST-Meta Learner} to obtain the node-level meta knowledge in both spatial and temporal domain. 
The non-shared feature extractor parameters $\theta_{ST}$ are generated to customize the feature extraction among source cities data and target city data.
\emph{ST-Meta Graph Reconstruction} is further designed for structure-aware meta training by reconstucting structure relations of different cities. The end-to-end learning process of ST-GFSL follows the MAML-based~\cite{DBLP:conf/icml/FinnAL17} episode learning.
%%H.Y.2.07: explain what is maml-based episode learning.
By mimicking few-shot scenarios in the target city, batches of few-shot training tasks are sampled to obtain a base model with strong adaptability.

\subsection{Spatio-Temporal Neural Network}

The Spatio-Temporal Neural Networks (STNN) can be divided into feature extractor and multi-step predictor, as shown in the bottom dotted box of Figure \ref{fig:framework}. The multi-step predictors often use one or more layers of fully connected networks ~\cite{DBLP:conf/ijcai/WuPLJZ19, DBLP:conf/cikm/LuGJFZ20,DBLP:conf/ijcai/YuYZ18} in literature. The feature extractor is designed according to different tasks and data characteristics, like RNN-based, CNN-based and GNN-based models. 
For example, in the experiment, we select the classical time series analysis networks GRU~\cite{DBLP:journals/corr/ChungGCB14} and TCN~\cite{DBLP:conf/cvpr/LeaFVRH17}, as well as the superior spatio-temporal graph neural network models STGCN~\cite{DBLP:conf/ijcai/YuYZ18} and GWN~\cite{DBLP:conf/ijcai/WuPLJZ19}.
Since ST-GFSL is designed for a model-agnostic framework, the parameter generation method adaptively generates non-shared feature extractor parameters $\theta_{ST}$ according to corresponding model structure. Therefore, our proposed ST-GFSL enables to benefit data-scarce scenarios from recent technique breakthroughs in spatio-temporal graph learning.

\subsection{ST-Meta Learner}

% Therefore, we propose spatio-temporal graph meta knowledge transfer mechanism. Instead of using parameter-shared feature extractor, we generate non-shared model parameters according to node-level spatio-temporal meta knowledge. 
% Specifically, we first design the ST-Meta Learner to extract spatio-temporal meta knowledge $\mathbf{Z}^{MK}$ and construct ST-meta graph $A_{meta}$. Afterwards, for different network structures of feature extractor, we design corresponding parameter generation strategies to generate node-level non-shared parameters.
%  \begin{figure}
%     \centering
%     \includegraphics[width=\linewidth]{figures/stmklearner.pdf}
%     \caption{Structure of ST-Meta Learner.}
%     \label{fig:mklearner}
% \end{figure}
 
% \subsubsection{ST-Meta Learner}

% GRU derives the representations of different hidden state, which is expressed as:
Spatio-Temporal Meta Knowledge Learner (ST-Meta Learner) extracts the node-level meta knowledge in both spatial and temporal domains. 
To encode the temporal dynamics of spatio-temporal graph, we employ Gated Recurrent Unit (GRU) ~\cite{DBLP:journals/corr/ChungGCB14}, which is widely used in time series modeling. Compared with classical RNN model, GRU retains the ability to extract long sequences and reduces the problem of gradient vanishing or exploding. 
Take node $v_i$ as an example, the node-level temporal meta knowledge $z_{i}^{tp}$ is expressed as the final state of $h_{i,t}$:
\begin{equation}
    \begin{aligned}
    z &=\sigma\left(U^{z} x_{i, t} + W^{z} h_{i, t-1} \right) \\
    r &=\sigma\left(U^{r} x_{i, t} + W^{r}h_{i, t-1} \right) \\
    c &=\phi \left(U^{c} x_{i, t} +W^{c} \left(h_{i, t-1} \circ r\right) \right) \\
    h_{i, t} &=(1-z) \circ c+z \circ h_{i, t-1}, 
    \end{aligned}
\end{equation}
where $x_{i, t} \in \mathbb{R}^d$ is the input vector of node $v_i$ at time $t$, and $h_{i, t-1}$ is the hidden state at time $t-1$. $U^{z}, U^{r}, U^{c} \in \mathbb{R}^{d\times d^{\prime}}$ and $W^{z}, W^{r}, W^{c} \in \mathbb{R}^{d^{\prime}\times d^{\prime}}$ are weight matrices. $\circ$ is the element-wise multiplication, $\sigma$ is the nonlinear activation function \emph{sigmoid}, and $\phi$ is \emph{tanh}. Thus, we derive the temporal meta knowledge of one city with GRU model, denoted as $\mathbf{Z}^{tp} = (z_1^{tp}, z_2^{tp}, \cdots, z_N^{tp}) \in \mathbb{R}^{N \times d^{\prime}}$. 

To encode the spatial correlations of spatio-temporal graph, we utilize spatial-based graph attention network (GAT)~\cite{DBLP:conf/iclr/VelickovicCCRLB18} for feature extraction. Graph attention is treated as a message-passing process in which information can be passed from one node to another along edges directly.
% due to the dynamic correlations of different node neighbors, 
% GAT is capable of capturing dynamic correlations of different nodes.
%%HY.2.7: This sentence is not precise. GCN is also a spectral-based method, which can be easily adapt to various structures.
Compared with spectral-based graph neural network~\cite{DBLP:conf/nips/DefferrardBV16, DBLP:conf/iclr/XuSCQC19}, GAT can adapt to various network structures. 
Therefore, it is suitable to learn spatial meta knowledge among multiple datasets. 

Specifically, we first apply a shared linear transformation to each group of interconnected nodes and compute the attention coefficients $e_{ij}$:
\begin{equation}
    e_{ij} = attention(W h_i, W h_j), j\in \mathcal{N}_{i} \text{,}
\end{equation}
where $W \in \mathbb{R}^{d \times O}$ is the weight matrix and $\mathcal{N}_{i}$ is a set of neighbor nodes of node $v_{i}$. The attention mechanism makes $\mathbb{R}^{O} \times \mathbb{R}^{O}\rightarrow \mathbb{R}$. After that, the attention score is normalized across all choices of $j$ using the $softmax$ function:
\begin{equation}
    \alpha_{ij} = \mathop{softmax}\nolimits_{j} (e_{ij}) = \frac{exp(e_{ij})}{\sum_{k \in \mathcal{N}_{i}}exp(e_{ik})} \text{.}
\end{equation}
In order to obtain more abundant representation, we execute the attention mechanism for $K$ times independently and employ averaging to achieve the spatial meta knowledge of node $v_i$:
\begin{equation}
    z_{i}^{sp} = \sigma(\frac{1}{K}\sum\nolimits_{k=1}^{K}\sum\nolimits_{j\in \mathcal{N}_{i}}\alpha_{ij}W^{k}h_{j})\text{.}
\end{equation}
Thus, we derive the spatial meta knowledge of one city $\mathbf{Z}^{sp} = (z_1^{sp}, z_2^{sp}, \cdots, z_N^{sp}) \in \mathbb{R}^{N \times d^{\prime}}$. 

%%HY.2.7: how to integrate, be more specific
By integrating spatio-temporal features, we obtain the meta knowledge denoted as $\mathbf{Z}^{MK}=(z_1^{MK}, z_2^{MK}, \cdots, z_N^{MK}) \in \mathbb{R}^{N \times d_{MK}}$. We weighted-sum the temporal and spatial meta knowledge through a learnable ratio $\gamma$, and $\gamma \in \mathbb{R}^{d^{\prime}}$. It learns the impact from spatial or temporal domain in a data-driven manner. Compared with the classical concatenation method, it is easier to adapt the spatio-temporal characteristics of cross city data through a learnable ratio. Meanwhile, it reduces the amount of parameters for data generation, which will be further discussed in \emph{Parameter Generation}. To be specific, the meta knowledge $\mathbf{Z}^{MK}$ is calculated as follows:
\begin{equation}
    \mathbf{Z}^{MK} = W^{\gamma}(\gamma \circ \mathbf{Z}^{tp} + (1-\gamma) \circ \mathbf{Z}^{sp}) \text{,}
\end{equation}
where $W^{\gamma} \in \mathbb{R}^{d^{\prime} \times d_{MK}}$ is the weight matrix for meta knowledge output layer, and $d_{MK}$ is the dimension of meta knowledge.

% meta knowledge is utilized to reconstruct the ST-Meta Graph for structure-aware learning and generate non-shared model parameters for spatio-temporal feature extraction.

\subsection{ST-Meta Graph Reconstruction}

In order to express the structural information of graphs and reduces structure deviation caused by different source data distribution, ST-Meta Graph is reconstructed by meta knowledge for structure-aware learning. We predict the likelihood of an edge existing between nodes $v_i$ and $v_j$, by multiplying learned meta knowledge $z_i^{MK}$ and $z_j^{MK}$ as follows:
\begin{equation}
    p(a_{ij} | z_i^{MK},z_j^{MK}) = sigmoid((z_i^{MK})^T, z_j^{MK}).
\end{equation}
As such, the ST-meta graph $\mathbf{A}_{meta}$ can be constructed as
\begin{equation}
    \mathbf{A}_{meta} = sigmoid[(\mathbf{Z^{MK}})^T \cdot {\mathbf{Z^{MK}}}],
\end{equation}
where $(\cdot)^T$ is the transpose of meta knowledge matrix. 

In order to guide the structure-aware learning of meta knowledge, we introduce \emph{graph reconstruction loss} $\mathcal{L}_g$ between the ST-meta graph $A_{meta}$ and the original adjacency matrix $A$ in training process, which is calculated as follows:
\begin{equation}
    \label{Lg}
    \mathcal{L}_g = \| \mathbf{A}_{meta} - \mathbf{A} \|^2.
\end{equation}

% For the training of meta knowledge representation, on one hand, it is end-to-end optimized through back propagation of the error loss function between the prediction value and the ground truth. On the other hand, we hope that the meta knowledge can fully express the structural information of graph and reduce feature deviation caused by different source data distribution. Therefore, we introduce \emph{ST-Meta Graph} in learning process. 
% We can predict the likelihood of an edge existing between two nodes, $v_i$ and $v_j$, by multiplying the learned meta knowledge of two nodes as follows:
% \begin{equation}
%     p(a_{ij} | z_i,z_j) = sigmoid(z_i^T, z_j)
% \end{equation}
% Therefore, the ST-meta graph $A_{meta}$ can be constructed as
% \begin{equation}
%     \mathbf{A}_{meta} = sigmoid[(\mathbf{Z^{MK}})^T \cdot {\mathbf{Z^{MK}}}]
% \end{equation}
% where $T$ is the transpose of meta knowledge matrix. 
% ST-Meta graph represents the structure information based on node-level spatio-temporal meta knowledge. The graph 

\subsection{Parameter Generation}

% Since different cities have different temporal and spatial characteristics, shallow models cannot effectively learn the complex features among multiple cities, and under-fitting problem occurs. Meanwhile, deep models are prone to overfitting during the adaptation to the target domain with few-shot data. Therefore, we propose parameter generation to obtain the non-shared parameters of feature extractors according to different node input features. The nodes in target city transfer the knowledge of source cities via parameter matching, retrieving from similar spatio-temporal characteristics.

After we obtain the node-level meta knowledge, due to the great difference across cities and within cities, we propose parameter generation to obtain the non-shared parameters of feature extractors for different scenarios.
When the meta-knowledge of a node in target domain is similar to that of a node in multiple source domains, approximate model parameters will be obtained. 
In other words, the nodes in target city transfer the knowledge of source cities via parameter matching, retrieving from similar spatio-temporal characteristics.

% \weinan{I think how you retrieve `similar' nodes is quite intersting but you seems not to include the details here.}
% lubin comments: i have revised and added the details about retrieving similar spatio-temporal features for parameter matching in above paragraph.

Our parameter generation is a function $\mathcal{F}$ that takes node-level meta knowledge as input and outputs the non-shared feature extractor parameters $\theta_{ST}$. Specifically, \emph{linear layer} and \emph{convolutional layer} are two basic neural network units. The following introduces how to generate the parameters of these two units respectively.

% Since different cities have different temporal and spatial characteristics, simple models cannot effectively learn the complex features, and under-fitting problem occurs. At the same time, complex models are prone to be overfitting during the adaptation to the target domain with few-shot data. Therefore, the parameter generation method can obtain the parameters of feature extractors according to different node input features, and achieve the balance between the source domain and the target domain.
% In ST-GFSL, different model parameters can be generated according to the structure of feature extractor.

\emph{Linear Layer} \, The expression of the linear layer is $\mathbf{Y}=\mathbf{WX}+\mathbf{b}$, where $\mathbf{W} \in \mathbb{R}^{d_{out}\times d_{in}}$ is the weight matrix and $\mathbf{b} \in \mathbb{R}^{d_{out}}$ is the bias. We can generate the model parameters $\mathbf{W}$ and $\mathbf{b}$ based on meta knowledge $\mathbf{Z}^{MK}$.

\begin{figure}[h]
    \centering
    \includegraphics[width=\linewidth]{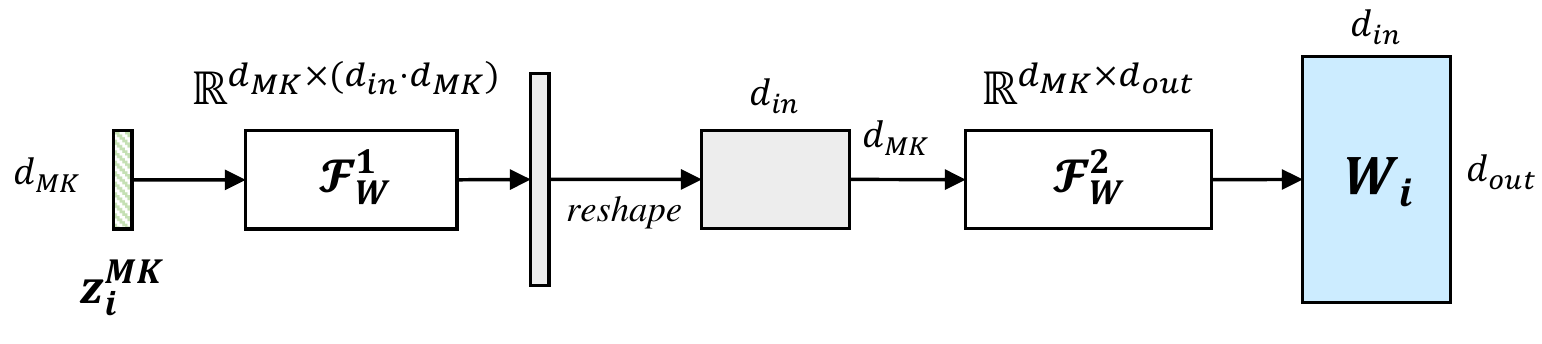}
    \caption{Parameter generation of linear weight matrix $W$.}
    \label{fig:param_w}
\end{figure}

Take node $v_i$ as an example, based on its meta knowledge $z_i^{MK} \in \mathbb{R}^{d_{MK}}$, the non-shared weight matrix $\mathbf{W_i}$ is generated through a two-step method as shown in Figure \ref{fig:param_w}. 
First, we perform a linear transformtion through $\mathcal{F}_W^1: \mathbb{R}^{d_{MK}} \rightarrow \mathbb{R}^{{d_{in} \cdot d_{MK}}}$ and conduct a dimension transformation: $\mathbb{R}^{{d_{in} \cdot d_{MK}}} \rightarrow \mathbb{R}^{d_{in} \times d_{MK}}$. Secondly, we perform the second linear transformation $\mathcal{F}_W^2: \mathbb{R}^{d_{in} \times d_{MK}} \rightarrow \mathbb{R}^{d_{in} \times d_{out}}$ and achieve the weight matrix $\mathbf{W_i}$ of node $v_i$. The parameter generation of bias $\mathbf{b}$ can be obtained directly by once linear transformation $\mathcal{F}_b$: $\mathbb{R}^{d_{MK}} \rightarrow \mathbb{R}^{d_{out}}$. 

The parameter number of the weight matrix $\mathbf{W_{i}}$ generated by two-step methods is $d_{MK} \cdot [d_{in} \times d_{MK} + d_{out}]$. Compared with direct generation of one linear layer ($\mathbb{R}^{d_{MK}} \rightarrow \mathbb{R}^{d_{in} \times d_{out}}$, the parameter number is $d_{MK} \times d_{in} \times d_{out}$). Obviously, two-step generation has less parameters. For example, in our experiment, when $d_{MK}=16$, $d_{in}=8$, $d_{out}=32$, the parameter number of using the two-step method is 2560. Whereas, the parameter number of directly using the one-step method is 4096. Our parameter generation method reduces the number of parameters by 37.5\%.

\begin{figure}[h]
    \centering
    \includegraphics[width=\linewidth]{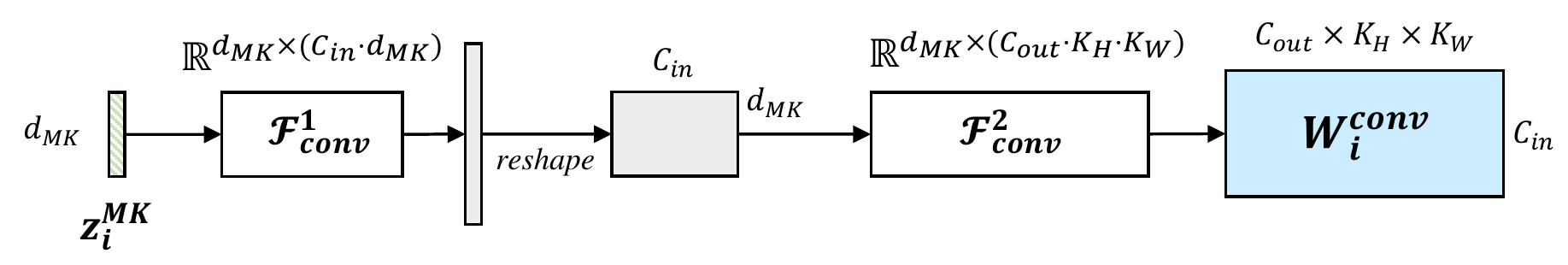}
    \caption{Parameter generation of 2D-conv weight matrix.}
    \label{fig:param_conv}
\end{figure}

\emph{Convolutional Layer} \,
Similar to the linear layer, the parameter generation of the convolutional layer adopts the two-step method shown in Figure \ref{fig:param_conv}, where $C_{in}$ is the number of channels of input data, $C_{out}$ is the number of channels of output data, and $(K_{H}, K_{W})$ is the size of the convolution kernel. Through the two-step generation, the convolutional kernel $\mathbf{W}_{i}^{conv}$ of node $v_i$ can be obtained after dimension reshape: $\mathbb{R}^{C_{in} \times (C_{out} \cdot K_H \cdot K_W)} \rightarrow \mathbb{R}^{C_{in} \times C_{out} \times K_H \times K_W}$. For 1D-convolution, the model parameters can be obtained only by adjusting the dimension of the convolutional kernel.

%%HY.2.7: Do you reduce the number of parameters in convolutional layers? If yes, you can point it out.

\subsection{ST-GFSL Learning Process}

To handle adaptation with few-shot scenarios, the learning process of ST-GFSL follows the MAML-based episode learning process. ST-GFSL trains the spatio-temporal graph learning model with two stage: \emph{base-model meta training} and \emph{adaptation}. In the \emph{base-model meta training} stage, inspired by MAML~\cite{DBLP:conf/icml/FinnAL17}, ST-GFSL imitates the adaptation process when encountering a new few-shot scene, and optimizes the adaptation capability.
Different graph sequences are sampled from multiple large-scale datasets (source datasets) to form a batch of training tasks $\mathcal{T}_{ST}$. Each task $\mathcal{T}_i \in \mathcal{T}_{ST}$ includes $K_\mathcal{S}$ support sets $\mathcal{S}_{i}$ and $K_\mathcal{Q}$ query sets $\mathcal{Q}_{i}$. In the \emph{adaptation} stage, the ST-GFSL model updates the parameters via several gradient descent steps on target domain data.

% \begin{figure}
%     \centering
%     \includegraphics[width=0.8\linewidth]{figures/framework_meta_train.pdf}
%     \caption{ST-GFSL Meta Training Stage Framework.}
%     \label{fig:st_maml}
% \end{figure}

To be specific, ST-GFSL first samples batches of task sets $\mathcal{T}_{ST}$ from source datasets. Each task $\mathcal{T}_i \in \mathcal{T}_{ST}$ belongs to one single city, and
is divided into support set $\mathcal{S}_{\mathcal{T}_i}$, query set $\mathcal{Q}_{\mathcal{T}_i}$ and $\mathcal{S}_{\mathcal{T}_i} \cap \mathcal{Q}_{\mathcal{T}_i} = \emptyset$. 
When learning a task $\mathcal{T}_i$, ST-GFSL considers a joint loss function
combining prediction error loss $\mathcal{L}_e$ and graph reconstruction loss $\mathcal{L}_g$. The prediction error loss $\mathcal{L}_e$ is the root mean square error between multi-step prediction and ground truth of support set $\mathcal{S}_{\mathcal{T}_i}$:
\begin{equation}
\label{Le}
    \mathcal{L}_e = \frac{1}{\|\mathcal{S}_{\mathcal{T}_i}\|}\sum_{(x_j, y_j)\in \mathcal{S}_{\mathcal{T}_i}} (f_{\theta}(x_j) - y_j)^2.
\end{equation}

As given in Equation \ref{Lg}, graph reconstruction loss $\mathcal{L}_g$ represents the structure-aware capability of meta knowledge. 
% When learning a task $\mathcal{T}_i$, ST-GFSL considers a joint loss function
% combining graph reconstruction loss $\mathcal{L}_g$ and prediction error loss $\mathcal{L}_e$. As given in Equation \ref{Lg}, graph reconstruction loss represents the structure-aware capability of meta knowledge. 
% $\mathcal{L}_e$ is the root mean square error between multi-step prediction and ground truth of support set $\mathcal{S}_{\mathcal{T}_i}$:
% \begin{equation}
% \label{Le}
%     \mathcal{L}_e = \frac{1}{\|\mathcal{S}_{\mathcal{T}_i}\|}\sum_{(x_j, y_j)\in \mathcal{S}_{\mathcal{T}_i}} (f_{\theta}(x_j) - y_j)^2.
% \end{equation}
Consequently, the joint loss funtion $\mathcal{L}$ is:
% The graph reconstruction loss $\mathcal{L}_g$ between ST-Meta graph and input graph is calculated as Equation \ref{Lg}. 
% Combining previous graph reconstruction loss $\mathcal{L}_g$ in Equation \ref{Lg} and prediction error loss $\mathcal{L}_e$ in Equation \ref{Le}, we obtain the joint loss function $\mathcal{L}$:
\begin{equation}
    \label{loss_func}
    \mathcal{L} = \mathcal{L}_e + \lambda \mathcal{L}_g,
\end{equation}
where $\lambda$ is the sum scale factor of two loss functions. 
To be specific, the meta objective is to minimize the sum of task loss on query sets, which is expressed as follows:
\begin{equation}
    \theta^{*} = \mathop{\arg\min}_{\theta} \sum\limits_{\mathcal{T}_i \in \mathcal{T}_{ST}}\mathcal{L}_{\mathcal{T}_i}(f_{\theta^{\prime}_i}) .
\end{equation}
In order to achieve the optimal model parameter $\theta^{*}$, Algorithm \ref{algorithm} outlines the \emph{base-model meta training} process of ST-GFSL. First of all, we iteratively sample a batch of task sets $\mathcal{T}_{ST}$ from source datasets (line 4). With regard to training task $\mathcal{T}_i \in \mathcal{T}_{ST}$, task-specific model parameter $\theta_{\mathcal{T}_i}^{\prime}$ is updated by gradient descents for several steps (line 8-9). The gradient of $f_{\theta_i^{\prime}}$ based on query set $Q_{\mathcal{T}_i}$ is given in line 10. Finally, the general model parameters $\theta$ are trained based on the summation across all the meta-training tasks (line 11).

\begin{algorithm}
  \KwIn{Source spatio-temporal Graph Dataset $\mathcal{G}_{ST}$, learning rate hyperparameter $\alpha, \beta$}
  
  \KwOut{Trained ST-GFSL model parameters $\theta^{*}$}
  $\theta \leftarrow$ random initalization\; // $\theta=\theta_{ST}+\theta_{\text{predictor}}$
  
  \While{not done}{
        Sample a batch of tasks $\mathcal{T}_{ST} \leftarrow \textsc{SampleTask} (\mathcal{G}_{ST})$ \;
        
        \For{$\mathcal{T}_i \in \mathcal{T}_{ST}$}{
            $\mathcal{S}_{\mathcal{T}_i} \leftarrow K_\mathcal{S}$ support set sample from $\mathcal{T}_i$ \;
            
            $\mathcal{Q}_{\mathcal{T}_i} \leftarrow K_\mathcal{Q}$ query set sample from $\mathcal{T}_i$ \;
            
            Evaluate $\nabla_\theta \mathcal{L}_{\mathcal{T}_i}(f_{\theta})$ with $S_{\mathcal{T}_i}$ via Equation (\ref{loss_func}) \;
            
            Compute adapted parameters with gradient descent: $\theta^{\prime}_{\mathcal{T}_i} = \theta - \alpha \nabla_\theta \mathcal{L}_{\mathcal{T}_i}(f_{\theta})$ \;
            
            Evaluate $\nabla_{\theta}\mathcal{L}_{\mathcal{T}_i}(f(\theta^{\prime}_{\mathcal{T}_i}))$ with $\mathcal{Q}_{\mathcal{T}_i}$ via Equation (\ref{loss_func}) \;
        }
        Update $\theta^{*} \leftarrow \theta - \beta \nabla_{\theta} \Sigma_{\mathcal{T}_i}\mathcal{L}_{\mathcal{T}_i}(f_{\theta^{\prime}_i})$
  }
  \caption{ST-GFSL base-model meta training}
  \label{algorithm}
\end{algorithm}

\section{Experiment}

In this section, we evaluate ST-GFSL in various aspects through extensive experiments. Specifically, we try to answer the following research questions through our evaluation:
\begin{itemize}
    \item [\textbf{RQ1}] How well does ST-GFSL perform against other baseline methods on different datasets?
\item [\textbf{RQ2}] How well are different spatio-temporal prediction models adaptable under the ST-GFSL framework?
\item [\textbf{RQ3}] How does each proposed component of model contribute to the performance of ST-GFSL?
\item [\textbf{RQ4}] How does each major hyperparameter affect the performance of ST-GFSL?
\end{itemize}
\subsection{Experiment Settings}
\subsubsection{Dataset} In the experiment, we take traffic flow prediction as an example to verify our proposed framework. We evaluate the performance of ST-GFSL on four traffic flow datasets: \emph{METR-LA}, \emph{PEMS-BAY}, \emph{Didi-Chengdu}, \emph{Didi-Shenzhen} ~\cite{DBLP:conf/iclr/LiYS018, Didi_dataset}.

In the experiment, the datasets are divided into source datasets, target dataset and test dataset. Take METR-LA as an example. When it is set as the target city, three-day data (a very small amount of data, compared with most works requiring several months of data) are selected as target data for \emph{adaptation} and the rest are regarded as test data. Other three datasets (PEMS-BAY, Didi-Shenzhen and Didi-Chengdu) are used as source datasets for \emph{meta training}. The same division method is used for other datasets and Z-score normalization are applied for data preprocessing.

\subsubsection{Metrics}
In order to fully verify the performance of our framework, we predict the traffic flow in the next 6 time steps with 12 historical time steps. Accordingly, the time step of METR-LA and PEMS-BAY datasets are 5 minutes, while Didi-Chengdu and Didi-Shenzhen datasets are 10 minutes due to the availability. Two widely used metrics are applied between the multi-step prediction and the ground truth for evaluation: Mean Absolute Error (MAE), and Root Mean Squared Error (RMSE).
% \begin{itemize}[left=1em]
%     \item Mean Absolute Error (MAE)
%         \begin{equation*}
%             MAE(y,\hat{y}) = \frac{1}{N}\sum\nolimits_{i=1}^{N}\left|\hat{y_{i}}-y_{i} \right|\text{.}
%         \end{equation*}
%     % \item Mean Absolute Percentage Error (MAPE)
%     %     \begin{equation*}
%     %         MAPE(y,\hat{y}) = \frac{100\%}{N}\sum\nolimits_{i=1}^{N}\left|\frac{\hat{y_{i}}-y_{i}}{y_{i}} \right|\text{.}
%     %     \end{equation*}
%     \item Root Mean Squared Error (RMSE)
%         \begin{equation*}
%             RMSE(y,\hat{y}) = \sqrt{\frac{1}{N}\sum\nolimits_{i=1}^{N}(\hat{y_{i}}-y_{i})^{2}}\text{.}
%         \end{equation*}
% \end{itemize}

% \subsubsection{Implementation} We implement ST-GFSL based on \emph{Pytorch} framework\footnote{The implementation code and details of our model is available at anonymous Github repo: https://github.com/RobinLu1209/ST-GFSL}. Totally, there are several important hyperparameters in our model, and we set them as: the dimension of meta knowledge $d_{MK} = 16$, task learning rate $\alpha = 0.01$, meta-training rate $\beta = 0.001$, task batch number $\|\mathcal{T}\| = 5$, and sum scale factor of two loss function $\lambda = 1.5$.

\begin{table*}[h]
\centering
\caption{Performance comparison of few-shot learning on four traffic speed datasets. In each column, the best results are highlighted in bold and grey, and the second best is underlined.}
\label{tab:performance}
\resizebox{\linewidth}{!}{%
\begin{tabular}{ccccccccccccc}
\hline
 &
  \multicolumn{6}{c}{\textbf{PEMS-BAY Dataset}} &
  \multicolumn{6}{c}{\textbf{METR-LA Dataset}} \\ \cline{2-13} 
 &
  \multicolumn{3}{c}{\textbf{MAE ($\downarrow$)}} &
  \multicolumn{3}{c}{\textbf{RMSE ($\downarrow$)}} &
  \multicolumn{3}{c}{\textbf{MAE ($\downarrow$)}} &
  \multicolumn{3}{c}{\textbf{RMSE ($\downarrow$)}} \\ \cline{2-13} 
\multirow{-3}{*}{\textbf{Baselines}} &
  \textbf{5 min} &
  \textbf{15 min} &
  \textbf{30 min} &
  \textbf{5 min} &
  \textbf{15 min} &
  \textbf{30 min} &
  \textbf{5 min} &
  \textbf{15 min} &
  \textbf{30 min} &
  \textbf{5 min} &
  \textbf{15 min} &
  \textbf{30 min} \\ \hline
HA &
  4.373 &
  4.373 &
  4.373 &
  6.745 &
  6.745 &
  6.745 &
  6.021 &
  6.021 &
  6.021 &
  9.483 &
  9.483 &
  9.483 \\
ARIMA &
  2.019 &
  2.307 &
  2.429 &
  3.929 &
  4.648 &
  5.360 &
  2.900 &
  3.058 &
  3.369 &
  4.179 &
  5.279 &
  7.670 \\
Target-only &
  1.556 &
  1.920 &
  2.368 &
  3.092 &
  4.043 &
  5.153 &
  2.740 &
  3.229 &
  3.860 &
  4.924 &
  6.118 &
  7.417 \\
Fine-tuned (Vanilla) &
  1.823 &
  2.166 &
  2.590 &
  3.434 &
  4.280 &
  5.276 &
  2.757 &
  3.277 &
  3.900 &
  4.883 &
  6.123 &
  7.413 \\
Fine-tuned (ST-Meta) &
  1.371 &
  1.791 &
  2.277 &
  2.699 &
  3.747 &
  4.920 &
  2.647 &
  3.188 &
  3.800 &
  4.368 &
  5.759 &
  7.110 \\
AdaRNN~\cite{Du2021ADARNN} &
  1.248 &
  1.928 &
  2.749 &
  2.084 &
  3.796 &
  5.725 &
  2.513 &
  {\ul 2.897} &
  \cellcolor[HTML]{EFEFEF}\textbf{3.312} &
  4.298 &
  \cellcolor[HTML]{EFEFEF}\textbf{5.567} &
  \cellcolor[HTML]{EFEFEF}\textbf{6.732} \\
MAML~\cite{DBLP:conf/icml/FinnAL17} &
  {\ul 1.081} &
  {\ul 1.600} &
  {\ul 2.141} &
  {\ul 1.906} &
  {\ul 3.291} &
  {\ul 4.708} &
  {\ul 2.405} &
  2.960 &
  3.639 &
  {\ul 4.159} &
  5.710 &
  7.124 \\ \hline
ST-GFSL (ours) &
  \cellcolor[HTML]{EFEFEF}\textbf{1.073} &
  \cellcolor[HTML]{EFEFEF}\textbf{1.560} &
  \cellcolor[HTML]{EFEFEF}\textbf{2.073} &
  \cellcolor[HTML]{EFEFEF}\textbf{1.865} &
  \cellcolor[HTML]{EFEFEF}\textbf{3.180} &
  \cellcolor[HTML]{EFEFEF}\textbf{4.584} &
  \cellcolor[HTML]{EFEFEF}\textbf{2.355} &
  \cellcolor[HTML]{EFEFEF}\textbf{2.896} &
  {\ul 3.557} &
  \cellcolor[HTML]{EFEFEF}\textbf{4.099} &
  {\ul 5.588} &
  {\ul 6.961} \\ \hline
\multicolumn{1}{l}{} &
  \multicolumn{1}{l}{} &
  \multicolumn{1}{l}{} &
  \multicolumn{1}{l}{} &
  \multicolumn{1}{l}{} &
  \multicolumn{1}{l}{} &
  \multicolumn{1}{l}{} &
  \multicolumn{1}{l}{} &
  \multicolumn{1}{l}{} &
  \multicolumn{1}{l}{} &
  \multicolumn{1}{l}{} &
  \multicolumn{1}{l}{} &
  \multicolumn{1}{l}{} \\ \hline
 &
  \multicolumn{6}{c}{\textbf{Didi-Chengdu Dataset}} &
  \multicolumn{6}{c}{\textbf{Didi-Shenzhen Dataset}} \\ \cline{2-13} 
 &
  \multicolumn{3}{c}{\textbf{MAE ($\downarrow$)}} &
  \multicolumn{3}{c}{\textbf{RMSE ($\downarrow$)}} &
  \multicolumn{3}{c}{\textbf{MAE ($\downarrow$)}} &
  \multicolumn{3}{c}{\textbf{RMSE ($\downarrow$)}} \\ \cline{2-13} 
\multirow{-3}{*}{\textbf{Baselines}} &
  \textbf{10 min} &
  \textbf{30 min} &
  \textbf{60 min} &
  \textbf{10 min} &
  \textbf{30 min} &
  \textbf{60 min} &
  \textbf{10 min} &
  \textbf{30 min} &
  \textbf{60 min} &
  \textbf{10 min} &
  \textbf{30 min} &
  \textbf{60 min} \\ \hline
HA &
  3.438 &
  3.438 &
  3.438 &
  4.879 &
  4.879 &
  4.879 &
  2.955 &
  2.955 &
  2.955 &
  4.342 &
  4.342 &
  4.342 \\
ARIMA &
  2.825 &
  3.305 &
  4.317 &
  3.889 &
  4.253 &
  5.597 &
  2.888 &
  3.056 &
  3.596 &
  4.489 &
  4.764 &
  5.575 \\
Target-only &
  2.386 &
  2.700 &
  3.085 &
  3.516 &
  4.017 &
  4.569 &
  2.071 &
  2.454 &
  2.834 &
  3.154 &
  3.793 &
  4.422 \\
Fine-tuned (Vanilla) &
  2.586 &
  2.877 &
  3.246 &
  3.746 &
  4.213 &
  4.751 &
  2.117 &
  2.490 &
  2.867 &
  3.196 &
  3.831 &
  4.442 \\
Fine-tuned (ST-Meta) &
  2.240 &
  2.693 &
  3.083 &
  3.249 &
  3.956 &
  4.519 &
  2.033 &
  2.454 &
  2.850 &
  2.989 &
  3.719 &
  4.385 \\
AdaRNN~\cite{Du2021ADARNN} &
  2.260 &
  2.724 &
  3.036 &
  3.231 &
  3.942 &
  \cellcolor[HTML]{EFEFEF}\textbf{4.324} &
  2.107 &
  2.473 &
  2.807 &
  3.041 &
  3.674 &
  4.231 \\
MAML~\cite{DBLP:conf/icml/FinnAL17} &
  {\ul 2.215} &
  {\ul 2.599} &
  {\ul 2.956} &
  {\ul 3.215} &
  {\ul 3.858} &
  4.399 &
  {\ul 1.917} &
  2.330 &
  {\ul 2.673} &
  {\ul 2.825} &
  {\ul 3.546} &
  {\ul 4.158} \\ \hline
ST-GFSL (ours) &
  \cellcolor[HTML]{EFEFEF}\textbf{2.188} &
  \cellcolor[HTML]{EFEFEF}\textbf{2.579} &
  \cellcolor[HTML]{EFEFEF}\textbf{2.927} &
  \cellcolor[HTML]{EFEFEF}\textbf{3.190} &
  \cellcolor[HTML]{EFEFEF}\textbf{3.820} &
  {\ul 4.339} &
  \cellcolor[HTML]{EFEFEF}\textbf{1.890} &
  \cellcolor[HTML]{EFEFEF}\textbf{2.288} &
  \cellcolor[HTML]{EFEFEF}\textbf{2.644} &
  \cellcolor[HTML]{EFEFEF}\textbf{2.763} &
  \cellcolor[HTML]{EFEFEF}\textbf{3.477} &
  \cellcolor[HTML]{EFEFEF}\textbf{4.100} \\ \hline
\end{tabular}%
}
\end{table*}

\subsection{Performance Comparison}

\subsubsection{Superiority of ST-GFSL over Baselines}

First, we compare the performance of ST-GFSL with a series of baselines on four datasets:
% [leftmargin=*]
\begin{itemize}[left=1em]
    \item \textbf{HA}: Historical Average, which formulates the traffic flow as a seasonal process, and uses average of previous seasons as the prediction.
    \item \textbf{ARIMA}: Auto-regressive integrated moving average is a well-known model that can understand and predict future values in a time series.
    \item \textbf{Target-only}: Directly training the model on few-shot data in target domain.
    \item \textbf{Fine-tuned (Vanilla)}: We first train the model on source datasets, and then fine-tune the model on few-shot data in target domain.
    \item \textbf{Fine-tuned (ST-Meta)}: Compared with ``Fine-tuned (Vanilla)'' method, we combine the proposed parameter generation based on meta knowledge to generate non-shared parameters for the model.
    \item \textbf{AdaRNN~\cite{Du2021ADARNN}}: A state-of-the-art transfer learning framework for non-stationary time series. This paper aims to reduce the distribution mismatch in the time series to learn an adaptive RNN-based model.
    \item \textbf{MAML~\cite{DBLP:conf/icml/FinnAL17}}: Model-Agnostic Meta 
    Learning (MAML), a superior meta-learning method that trains a model’s parameters such that a small number of gradient updates will lead to fast learning on a new task.
\end{itemize}

We can divide the above compared baselines into two categories: non-transfer methods and transfer methods. Non-transfer methods (HA, ARIMA, Target-only) only use the few-shot training data in target city. The rest are transfer methods, and these methods transfer the knowledge learned from multiple source datasets. Since AdaRNN uses GRU model for feature extraction, in order to maintain the fairness and consistency of comparison, other deep learning methods also use GRU model as feature extractor.

Table \ref{tab:performance} shows the performance comparison under various methods. We make the following observations: (1) Our proposed framework ST-GFSL obtains the best results on multiple datasets in both short-term and long-term predictions, demonstating the superiority of our method.
(2) In respect of AdaRNN, it performs better on some indicators, especially the mid- and long-term prediction on METR-LA dataset. The possible reason is that METR-LA is a relatively small-scale dataset with only 207 sensor nodes, and there is a strong temporal correlation between nodes. The AdaRNN's design for temporal covariate shift makes it have a better performance in long-term prediction. However, AdaRNN is unstable when adapting to other cities (e.g. PEMS-BAY), and the experiment performance fluctuates greatly on different datasets.
(3) For two fine-tuning methods, Fine-tuned (ST-Meta) based on parameter generation has a significant improvement compared to vanilla method. The results imply that our proposed non-shared parameters can better extract spatio-temporal features across the cities. 
(4) MAML shows good performance in the experiment. Meanwhile, MAML can be regarded as an ablation study of ST-GFSL, which will be further discussed in Section 5.3.

% MAML shows good performance in the experiment and the effectiveness of episode learning. At the same time, MAML can be regarded as an ablation experimental model after the degradation of our method, which will be introduced in Section 4.2.

% Secondly, our proposed framework ST-GFSL obtains the superior results on three datasets (PEMS-BAY, Didi-Shenzhen and Didi-Chengdu) in both short-term and long-term predictions, demonstating the benefits of meta knowledge transfer mechanism and structure-aware training process. In respect of the recently-proposed model, AdaRNN performs better on METR-LA dataset for mid- and long-term prediction. The possible reason is that METR-LA is a relatively small-scale dataset with only 207 sensor nodes, 
% and there is a strong temporal correlation between nodes. The AdaRNN design for temporal covariate shift makes it have a better performance in long-term prediction. However, AdaRNN is unstable when adapting to new cities (e.g. PEMS-BAY), and the experiment performance fluctuates greatly on different datasets. Lastly, MAML in performance comparison adds the graph reconstruction loss in meta-training process. Therefore, it can also be regarded as an ablation study of ST-GFSL and the details are further described in Section 5.3.

\subsubsection{ST-GFSL for different feature extractors}

ST-GFSL is a model-agnostic framework. In order to verify the versatility of the framework and further improve the prediction performance from STNN model aspect, we apply some advanced spatio-temporal data graph learning algorithms to our ST-GFSL framework. At the same time, we also train the following model directly on few-shot data in target domain to compare these two methods.

\begin{itemize}[left=1em]
    \item \textbf{TCN~\cite{DBLP:conf/cvpr/LeaFVRH17}}: 1D dilated convolution network-based temporal convolution network.
    \item \textbf{STGCN~\cite{DBLP:conf/ijcai/YuYZ18}}: Spatial temporal graph convolution network, which combines graph convolution with 1D convolution.
    \item \textbf{GWN~\cite{DBLP:conf/ijcai/WuPLJZ19}}: A convolution network structure combines graph convolution with dilated casual convolution and a self-adaptive graph.
\end{itemize}

\begin{figure}
    \centering
    \includegraphics[width=0.95\linewidth]{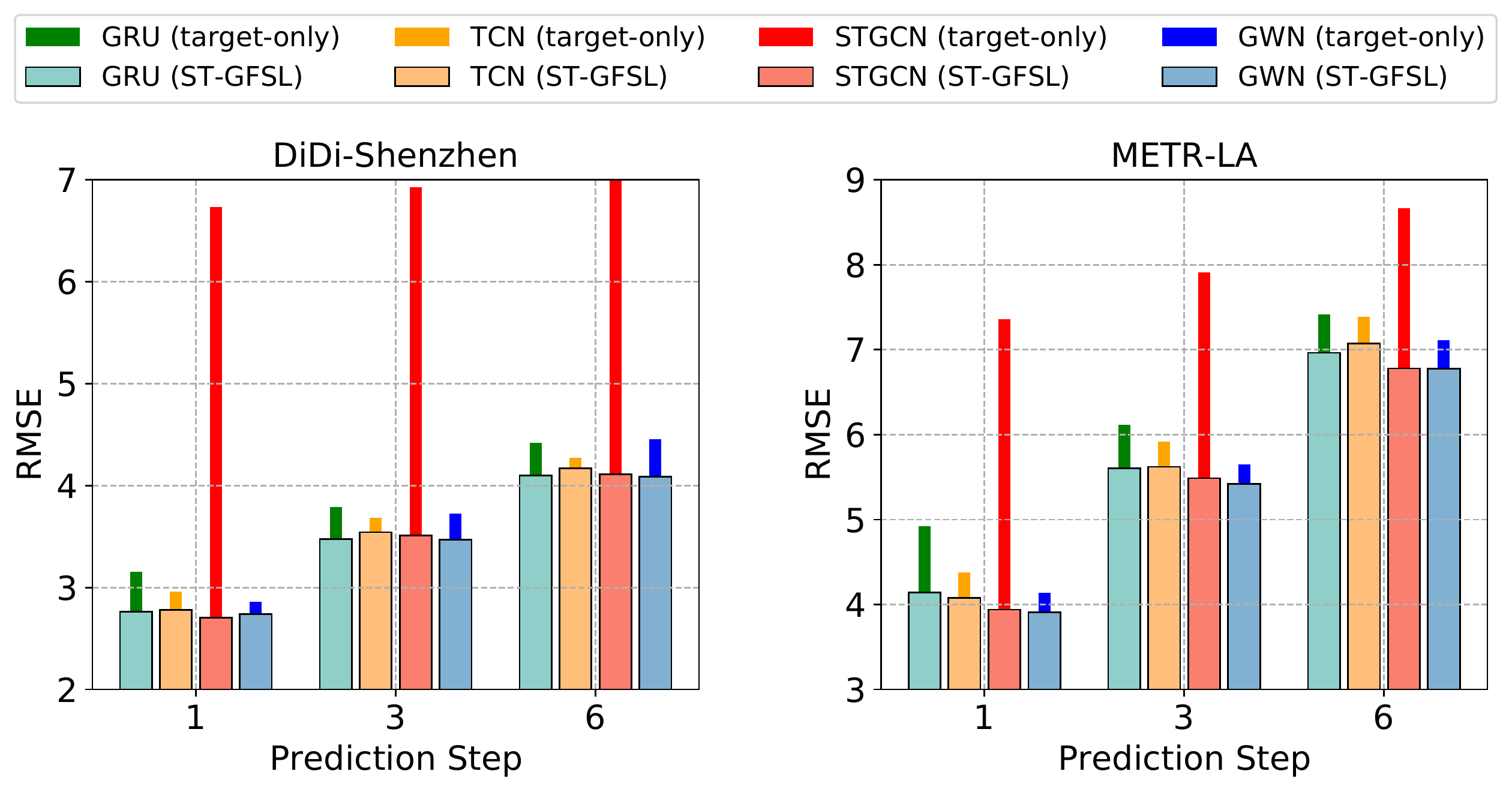}
    \caption{Performance comparison of different feature extractors on Didi-Shenzhen and METR-LA datasets.}
    \label{fig:model_compare}
\end{figure}
% The RMSE of the 6-th step prediction of GWN on Didi-Shenzhen dataset is reduced from 4.170 to 4.088 compared with TCN, and from 7.073 to 6.778 on METR-LA dataset.
Figure \ref{fig:model_compare} shows the performance of different feature extractors on Didi-Shenzhen and METR-LA datasets. STGCN and Graph WaveNet, as the representative work of traffic speed prediction, still maintain their strong feature extraction capabilities under the ST-GFSL framework. 
Compared with direct training on few-shot target domain, ST-GFSL improves the performance greatly. Especially for STGCN, direct training with few-shot samples has over-fitting, which makes it impossible to predict accurately.

% This shows that ST-GFSL is adapted to different network structures, such as RNN, CNN, GNN, etc., and has strong scalability and versatility.

% In view of the length of the paper, Figure \ref{fig:performance_diff_model} shows the performance of different base models on PEMS-BAY and Didi-Shenzhen datasets in the next several steps. It can be seen that the prediction error decreases with the upgrading of feature extractors. STGCN and Graph WaveNet, as the representative work of traffic speed prediction, still maintain their strong feature extraction capabilities under the ST-GFSL framework. This shows that ST-GFSL is adapted to different network structures, such as RNN, CNN, GNN, etc., and has strong scalability and versatility.

% \begin{figure}
%     \centering
%     \subfigure[Didi-Shenzhen Dataset]{
%         \includegraphics[width=\linewidth]{figures/performance_model_shenzhen.pdf}
%     }
%     \\
%     \subfigure[METR-LA Dataset]{
%         \includegraphics[width=\linewidth]{figures/performance_model_metr.pdf}
%     }
%     \caption{Performance comparison of different feature extractor with ST-GFSL on Didi-Shenzhen and METR-LA dataset.} 
%     \label{fig:performance_diff_model}
% \end{figure}
\subsubsection{Performance across the cities}
\begin{figure}
    \centering
    \subfigure[Didi-Shenzhen Dataset]{
        \includegraphics[width=\linewidth]{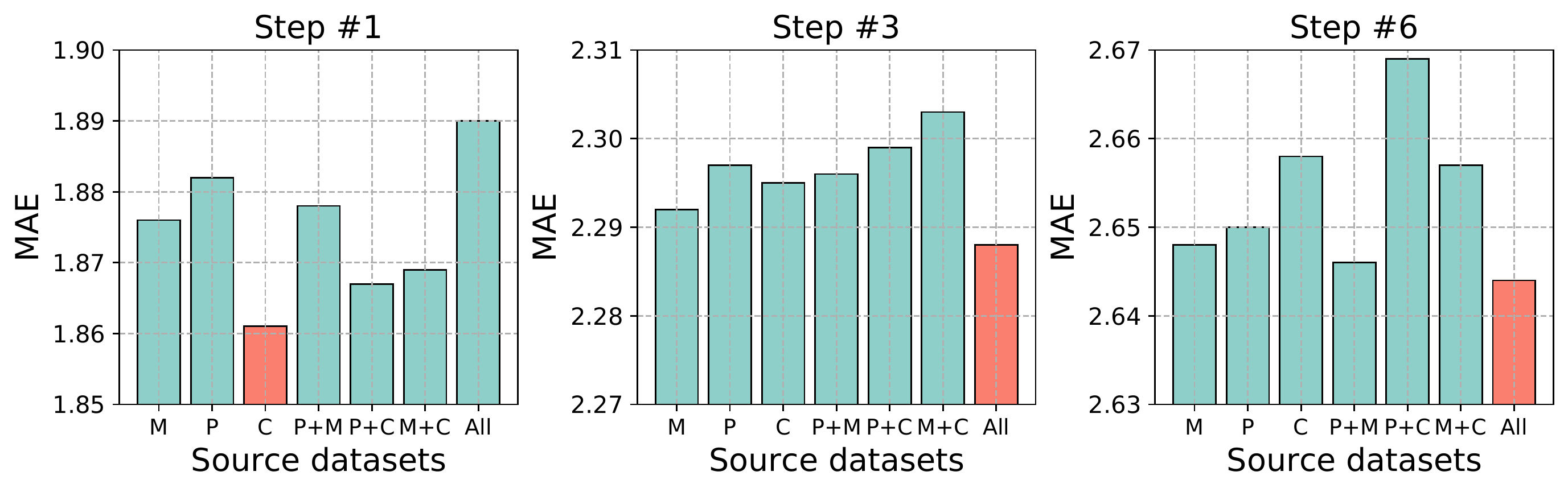}
    }
    \\
    \subfigure[METR-LA Dataset]{
        \includegraphics[width=\linewidth]{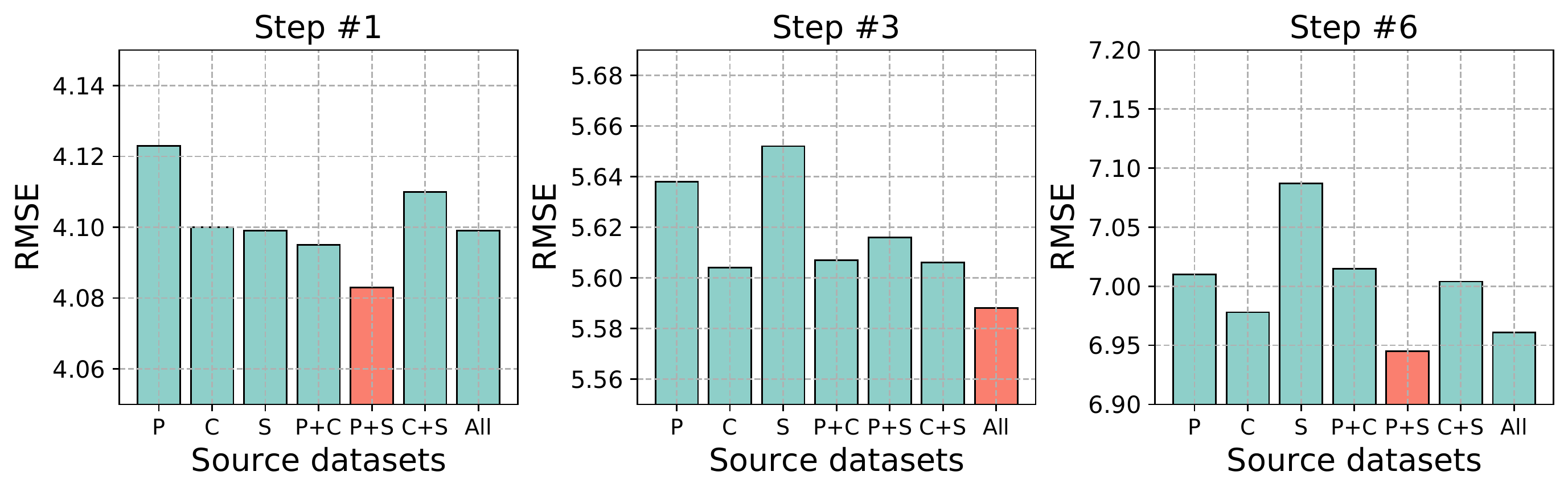}
    }
    \caption{Performance comparison across different source cities studied on Didi-Shenzhen and METR-LA dataset.} 
    \label{fig:source}
\end{figure}

In order to explore the performance of transfering across multiple cities, we conduct experiments on different source cities, and obtain the experimental results in Figure \ref{fig:source}. For simplicity, METR-LA dataset is denoted as ``M'', PEMS-BAY dataset is denoted as ``P'', Didi-Chengdu dataset is denoted as ``C'', and Didi-Shenzhen dataset is denoted as ``S''. 

As shown in Figure \ref{fig:source}(a), in Step \#1, only use Didi-Chengdu dataset as the source city obtain the best performance. In Step \#3 and Step \#6, the best results are obtained by transfering the knowledge of three cities. In the short-term prediction, since Shenzhen and Chengdu are both first tier cities in China, they have more similar spatio-temporal characteristics. In long-term prediction, the knowledge transfer of multiple cities can help better predict long-term correlations. For METR-LA dataset, better performance is usually achieved when ``P+S'' is used as the source domain. In Step \#3, the best performance can be obtained by using all datasets.

\begin{table*}[]
\centering
\caption{Ablation Studies of ST-GFSL on Didi-Shenzhen and METR-LA dataset.}
\label{tab:ablation}
\resizebox{\linewidth}{!}{%
\begin{tabular}{lllcccccccccccc}
\hline
\multicolumn{3}{l}{} &
  \multicolumn{6}{c}{\textbf{Didi-Shenzhen Dataset}} &
  \multicolumn{6}{c}{\textbf{METR-LA Dataset}} \\ \cline{4-15} 
\multicolumn{3}{l}{} &
  \multicolumn{3}{c}{\textbf{MAE ($\downarrow$)}} &
  \multicolumn{3}{c}{\textbf{RMSE ($\downarrow$)}} &
  \multicolumn{3}{c}{\textbf{MAE ($\downarrow$)}} &
  \multicolumn{3}{c}{\textbf{RMSE ($\downarrow$)}} \\ \cline{4-15} 
\multicolumn{3}{l}{\multirow{-3}{*}{\textbf{Ablation Methods}}} &
  \textbf{10 min} &
  \textbf{30 min} &
  \textbf{60 min} &
  \textbf{10 min} &
  \textbf{30 min} &
  \textbf{60 min} &
  \textbf{10 min} &
  \textbf{30 min} &
  \textbf{60 min} &
  \textbf{10 min} &
  \textbf{30 min} &
  \textbf{60 min} \\ \hline
\multicolumn{3}{l}{(M1a): Use temporal meta knowledge only} &
  1.910 &
  2.317 &
  2.668 &
  2.834 &
  3.530 &
  4.139 &
  \cellcolor[HTML]{EFEFEF}\textbf{2.332} &
  { 2.905} &
  3.616 &
  \cellcolor[HTML]{EFEFEF}\textbf{4.077} &
  5.619 &
  6.991 \\
\multicolumn{3}{l}{(M1b): Use spatial meta knowledge only} &
  { 1.872} &
  { 2.300} &
  { 2.649} &
  { 2.756} &
  { 3.495} &
  { 4.108} &
  { 2.364} &
  2.915 &
  { 3.604} &
  { 4.112} &
  { 5.612} &
  { 6.970} \\
\multicolumn{3}{l}{(M1c): Use random initialized vectors} &
  1.937 &
  2.332 &
  2.680 &
  2.841 &
  3.531 &
  4.142 &
  2.422 &
  2.949 &
  3.697 &
  4.182 &
  5.624 &
  7.047 \\
\multicolumn{3}{l}{(M2): Remove parameter generator} &
  1.917 &
  2.330 &
  2.673 &
  2.825 &
  3.546 &
  4.158 &
  2.405 &
  2.960 &
  3.639 &
  4.129 &
  5.655 &
  7.075 \\
\multicolumn{3}{l}{(M3): Remove graph reconstruction loss $\mathcal{L}_{g}$} &
  2.286 &
  2.652 &
  3.000 &
  3.309 &
  3.979 &
  4.604 &
  3.087 &
  3.585 &
  4.140 &
  4.960 &
  6.209 &
  7.464 \\ \hline
\multicolumn{3}{l}{\textbf{ST-GFSL (Ours)}} &
  \cellcolor[HTML]{EFEFEF}\textbf{1.856} &
  \cellcolor[HTML]{EFEFEF}\textbf{2.290} &
  \cellcolor[HTML]{EFEFEF}\textbf{2.634} &
  \cellcolor[HTML]{EFEFEF}\textbf{2.737} &
  \cellcolor[HTML]{EFEFEF}\textbf{3.471} &
  \cellcolor[HTML]{EFEFEF}\textbf{4.052} &
  2.387 &
  \cellcolor[HTML]{EFEFEF}\textbf{2.895} &
  \cellcolor[HTML]{EFEFEF}\textbf{3.546} &
  4.140 &
  \cellcolor[HTML]{EFEFEF}\textbf{5.603} &
  \cellcolor[HTML]{EFEFEF}\textbf{6.963} \\ \hline
\end{tabular}%
}
\end{table*}

\subsection{Ablation Study}

In this section, we verify the effectiveness of each module in ST-GFSL through ablation study. First of all, we only use time domain features (M1a) or only spatial domain features (M1b) as meta knowledge for parameter generation. 
We can see that the performance has decreased to a certain extent, which proves that the spatio-temporal joint features are more accurate for parameter generation. 
It is worth noting that the short-time prediction of METR-LA datasets achieves the best performance by using only temporal meta knowledge. This is because its temporal characteristics are more prominent.
Furthermore, we directly use trainable random parameters to replace the learned spatio-temporal features (M1c). Since the spatio-temporal attributes are not explicitly defined, it is difficult for the random initialized vector to capture the complex and dynamic characteristics, so the performance has been greatly reduced. 
When we remove the parameter generator directly (M2), it degenerates into a vanilla neural network model trained by proposed ST-GFSL framework. Compared with the non-shared model parameters generated by meta knowledge, the performance has more degradation. When we subtract the graph reconstruction loss function during the training process (M3), the model has a severe performance degradation. The reason lies in that meta knowledge is not only expected to be able to extract effective spatio-temporal features, but more importantly, these features are compatible with structural information and can be generalized to more scenarios.
% Due to space limitations, Table \ref{tab:ablation} shows the results on METR-LA and Didi-Chengdu datasets.The other two datasets also have similar results.

% For the spatio-temporal meta knowledge, first of all, we only use time domain features (M1a) or only spatial domain features (M1b) as meta knowledge for parameter generation. We can see that the model performance has decreased to a certain extent, which proves that the spatio-temporal joint features are more accurate for parameter generation. 
% % However, we see the advantages of several indicators in (M1a) and (M1b). This is because the temporal characteristics of METR-LA are more prominent, and the spatial characteristics of Didi-Chengdu are more dominant. Nevertheless, in general, the combination of temporal and spatial features can have a better effect. 
% In addition, we directly use trainable random parameters to replace the learned spatio-temporal features (M1c). Since the spatio-temporal attributes are not explicitly defined, it is difficult for the random initialized vector to capture the complex and dynamic characteristics, so the performance has been greatly reduced. 

% \usepackage{multirow}
% \usepackage{graphicx}

\subsection{Case Study}

\begin{figure}
    \centering
    \includegraphics[width=\linewidth]{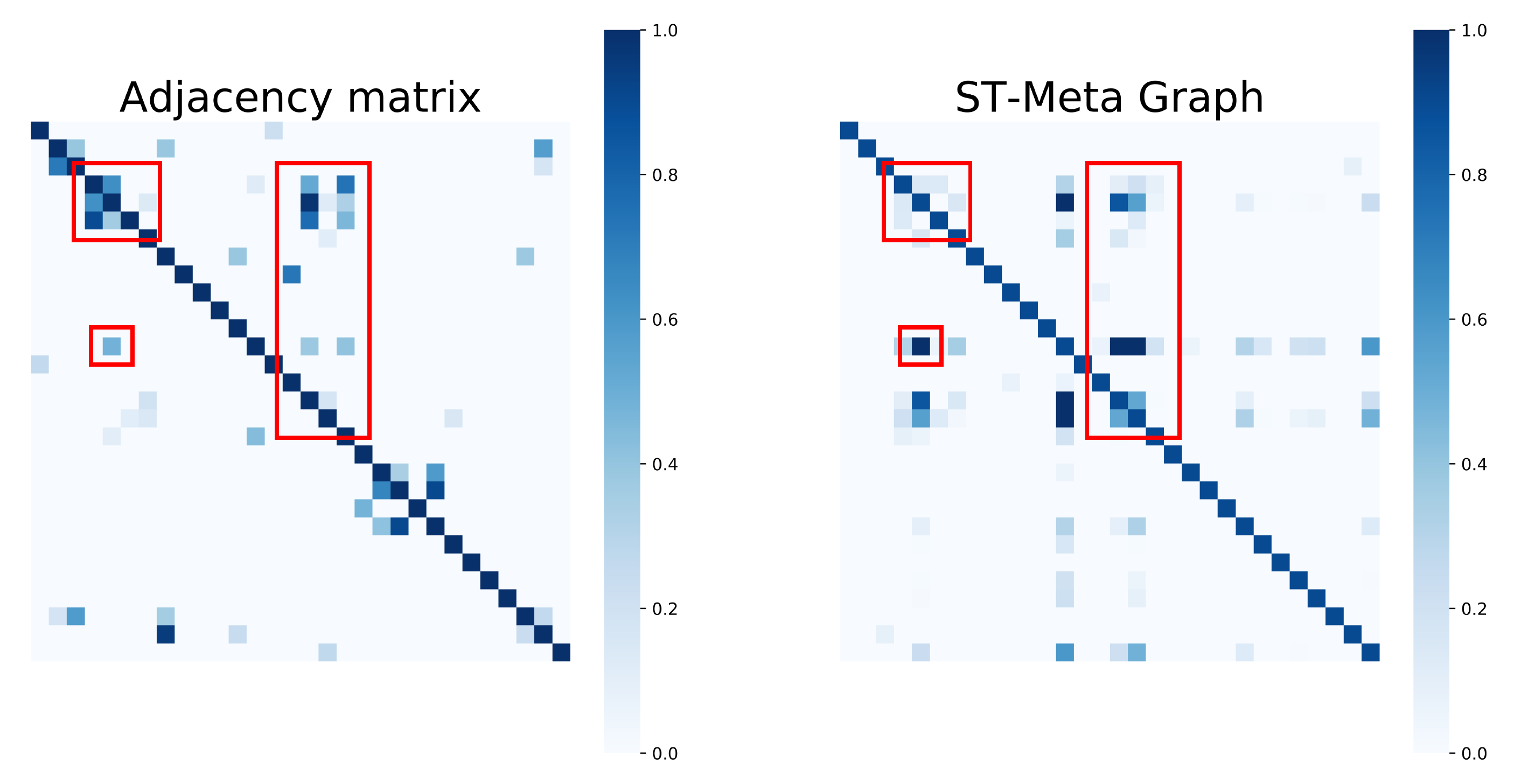}
    \caption{Case study on visualization of adjacency matrix and reconstructed ST-Meta graph.}
    \label{fig:case_study}
\end{figure}

In order to further explore the influence of \emph{graph reconstruction loss} in cross-city knowledge transfer, we perform visual analysis on the original adjacency matrix and reconstructed ST-Meta graph. For clarity, we select the adjacency relationship of the first 30 nodes as shown in Figure \ref{fig:case_study}. It can be seen that in the process of graph reconstruction, some key structural relationships (represented by red boxes in the figure) are better reconstructed. This shows that the proposed ST-meta graph reconstruction achieves expected effects, and it greatly improves the prediction performance by avoiding structural deviation as shown in the ablation study.
Indeed, in the process of knowledge transfer learning across multiple cities, some structural information has not been well captured. This is a great challenge in graph data transfer learning, i.e., how to avoid structure deviation among multiple source datasets, which will become our future work.

\subsection{Hyperparameter Analysis}

\begin{figure}[h]
    \centering
    \includegraphics[width=\linewidth]{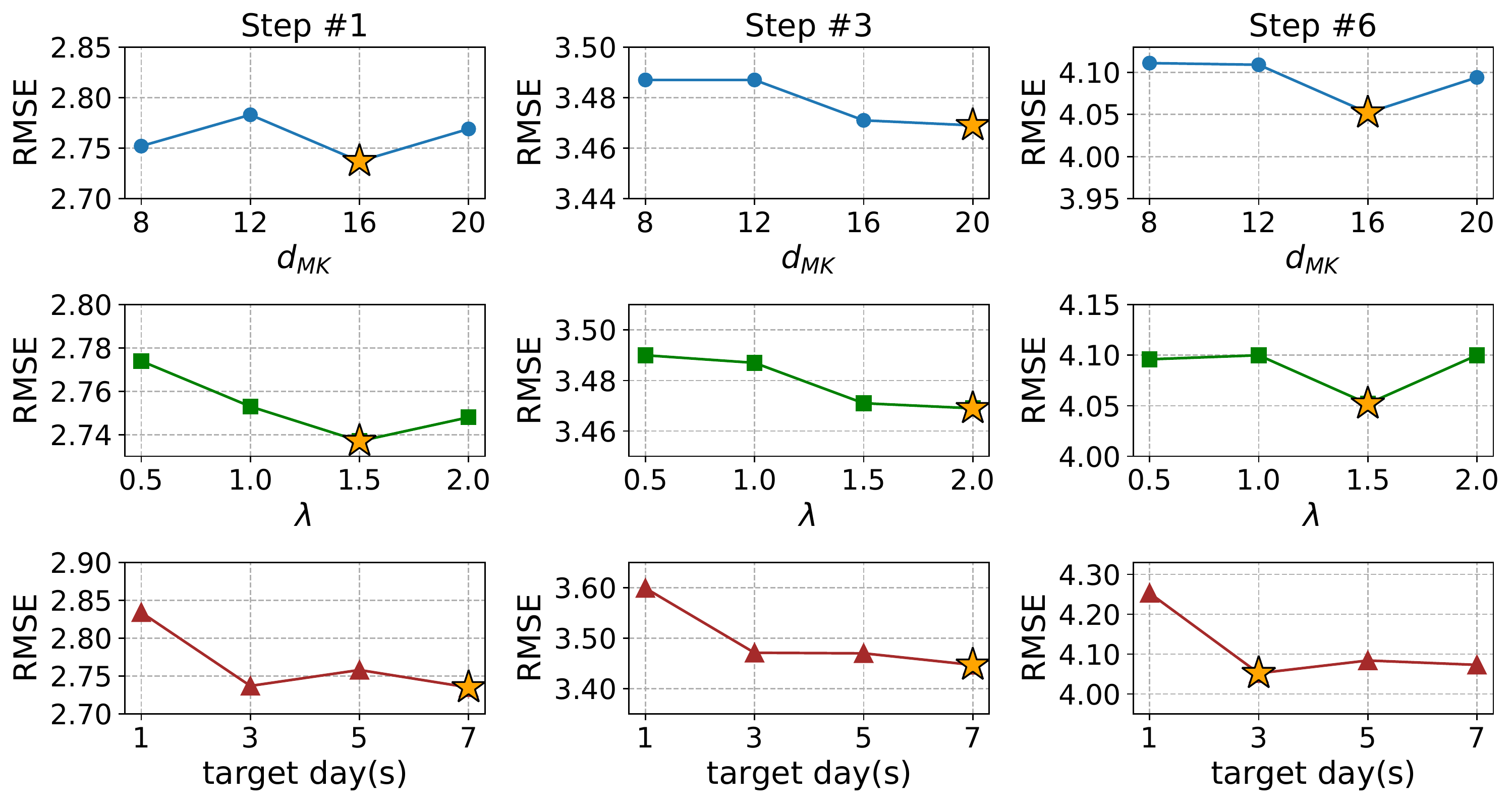}
    \caption{Hyperparameter study on Didi-Shenzhen dataset: meta knowledge dimension $d_{MK}$, sum scale factor $\lambda$ and training sample days of target city.}
    \label{fig:hyper}
\end{figure}

We select the hyperparameters by analyzing the experimental results as shown in Figure \ref{fig:hyper}. (1) First, we change the dimension of meta knowledge $d_{MK}$. The dimension of meta knowledge directly affects the efficiency of parameter generation. Overall, when $d_{MK}=16$, there are better performance results in both short and long term prediction. (2) Besides, the trade-off between the two loss functions is an important consideration. By observing the changes of two loss functions, we set $\lambda$ to change from 0.5 to 2.0 in the hyperparameter experiments. When we set $\lambda \textgreater 1$, we can often get better results, which also reflects the importance of graph reconstruction loss. (3) In the experiment, we all use 3-days target domain data as few-shot samples. When we adjust from 1 day to 7 days, the improvement of model performance is gradually significant in general.

\section{Conclusion}
In this paper, we first propose a spatio-temporal graph few-shot learning framework called ST-GFSL for cross-city knowledge transfer. 
Non-shared feature extractor parameters based on node-level meta knowledge improve the effectiveness of spatio-temporal representation on multiple datasets and transfer the cross-city knowledge via parameter matching from similar spatio-temporal meta knowledge.  
ST-GFSL integrates the graph reconstruction loss to achieve structure-aware learning.
Extensive experimental results in the running case of traffic speed prediction demonstrate the superiority of ST-GFSL over other baseline methods. Besides traffic speed prediction, ST-GFSL could be applied to other
few-shot scenarios equipped with spatio-temporal graph learning, such as taxi demand prediction in different cities, indoor environment monitoring in different warehouses, etc. In the future, we will further explore the problem of structure deviation in knowledge transfer among graph data.

% the knowledge transfer considering the heterogeneity of graphs.

\section*{Acknowledgement}
This work was partially supported by National Key R\&D Program of China (No.2018AAA0101200), NSF China (No. 42050105, 62020106005, 62061146002, 61960206002, 61829201, 61832013), the Science and Technology Innovation Program of Shanghai (No.19YF1424500), and the Program of Shanghai Academic/Technology Research Leader under Grant No.18XD1401800.

\bibliographystyle{unsrt}
\bibliography{reference}

\begin{thebibliography}{10}

\bibitem{DBLP:conf/ijcai/WuPLJZ19}
Zonghan Wu, Shirui Pan, Guodong Long, Jing Jiang, and Chengqi Zhang.
\newblock Graph wavenet for deep spatial-temporal graph modeling.
\newblock In {\em Proceedings of the Twenty-Eighth International Joint
  Conference on Artificial Intelligence, {IJCAI} 2019, Macao, China, August
  10-16, 2019}, pages 1907--1913, 2019.

\bibitem{DBLP:conf/cikm/LuGJFZ20}
Bin Lu, Xiaoying Gan, Haiming Jin, Luoyi Fu, and Haisong Zhang.
\newblock Spatiotemporal adaptive gated graph convolution network for urban
  traffic flow forecasting.
\newblock In {\em The 29th {ACM} International Conference on Information and
  Knowledge Management, Virtual Event, Ireland, October 19-23, 2020}, pages
  1025--1034. {ACM}, 2020.

\bibitem{DBLP:conf/aaai/YeSDF021}
Junchen Ye, Leilei Sun, Bowen Du, Yanjie Fu, and Hui Xiong.
\newblock Coupled layer-wise graph convolution for transportation demand
  prediction.
\newblock In {\em Thirty-Fifth {AAAI} Conference on Artificial Intelligence,
  {AAAI} 2021, Virtual Event, February 2-9, 2021}, pages 4617--4625, 2021.

\bibitem{10.1145/3331184.3331368}
Zhenlong Zhu, Ruixuan Li, Minghui Shan, Yuhua Li, Lu~Gao, Fei Wang, Jixing Xu,
  and Xiwu Gu.
\newblock Tdp: Personalized taxi demand prediction based on heterogeneous graph
  embedding.
\newblock In {\em Proceedings of the 42nd International ACM SIGIR Conference on
  Research and Development in Information Retrieval}, SIGIR'19, page
  1177–1180, New York, NY, USA, 2019.

\bibitem{DBLP:conf/wsdm/WangZZLY21}
Chunyang Wang, Yanmin Zhu, Tianzi Zang, Haobing Liu, and Jiadi Yu.
\newblock Modeling inter-station relationships with attentive temporal graph
  convolutional network for air quality prediction.
\newblock In {\em {WSDM} '21, The Fourteenth {ACM} International Conference on
  Web Search and Data Mining, Virtual Event, Israel, March 8-12, 2021}, pages
  616--634. {ACM}, 2021.

\bibitem{DBLP:conf/ijcai/WangGMLY19}
Leye Wang, Xu~Geng, Xiaojuan Ma, Feng Liu, and Qiang Yang.
\newblock Cross-city transfer learning for deep spatio-temporal prediction.
\newblock In {\em Proceedings of the Twenty-Eighth International Joint
  Conference on Artificial Intelligence, {IJCAI} 2019, Macao, China, August
  10-16, 2019}, pages 1893--1899, 2019.

\bibitem{10.1145/2939672.2939830}
Ying Wei, Yu~Zheng, and Qiang Yang.
\newblock Transfer knowledge between cities.
\newblock In {\em Proceedings of the 22nd ACM SIGKDD International Conference
  on Knowledge Discovery and Data Mining}, KDD '16, page 1905–1914, New York,
  NY, USA, 2016.

\bibitem{10.1145/3308558.3313577}
Huaxiu Yao, Yiding Liu, Ying Wei, Xianfeng Tang, and Zhenhui Li.
\newblock Learning from multiple cities: A meta-learning approach for
  spatial-temporal prediction.
\newblock In {\em The World Wide Web Conference}, WWW '19, page 2181–2191,
  New York, NY, USA, 2019.

\bibitem{moreira2013predicting}
Luis Moreira-Matias, Joao Gama, Michel Ferreira, Joao Mendes-Moreira, and Luis
  Damas.
\newblock Predicting taxi--passenger demand using streaming data.
\newblock {\em IEEE Transactions on Intelligent Transportation Systems},
  14(3):1393--1402, 2013.

\bibitem{DBLP:conf/ijcai/BaiYK0S19}
Lei Bai, Lina Yao, Salil~S. Kanhere, Xianzhi Wang, and Quan~Z. Sheng.
\newblock Stg2seq: Spatial-temporal graph to sequence model for multi-step
  passenger demand forecasting.
\newblock In {\em Proceedings of the Twenty-Eighth International Joint
  Conference on Artificial Intelligence, {IJCAI} 2019, Macao, China, August
  10-16, 2019}, pages 1981--1987, 2019.

\bibitem{do2020graph}
Tien~Huu Do, Evaggelia Tsiligianni, Xuening Qin, Jelle Hofman, Valerio~Panzica
  La~Manna, Wilfried Philips, and Nikos Deligiannis.
\newblock Graph-deep-learning-based inference of fine-grained air quality from
  mobile iot sensors.
\newblock {\em IEEE Internet of Things Journal}, 7(9):8943--8955, 2020.

\bibitem{DBLP:conf/kdd/PanLW00Z19}
Zheyi Pan, Yuxuan Liang, Weifeng Wang, Yong Yu, Yu~Zheng, and Junbo Zhang.
\newblock Urban traffic prediction from spatio-temporal data using deep meta
  learning.
\newblock In {\em Proceedings of the 25th {ACM} {SIGKDD} International
  Conference on Knowledge Discovery {\&} Data Mining, {KDD} 2019, Anchorage,
  AK, USA, August 4-8, 2019}, pages 1720--1730. {ACM}, 2019.

\bibitem{pan_tkde_2020}
Zheyi Pan, Wentao Zhang, Yuxuan Liang, Weinan Zhang, Yong Yu, Junbo Zhang, and
  Yu~Zheng.
\newblock Spatio-temporal meta learning for urban traffic prediction.
\newblock {\em IEEE Transactions on Knowledge and Data Engineering}, pages
  1--1, 2020.

\bibitem{DBLP:conf/icml/FinnAL17}
Chelsea Finn, Pieter Abbeel, and Sergey Levine.
\newblock Model-agnostic meta-learning for fast adaptation of deep networks.
\newblock In {\em Proceedings of the 34th International Conference on Machine
  Learning, {ICML} 2017, Sydney, NSW, Australia, 6-11 August 2017}, volume~70
  of {\em Proceedings of Machine Learning Research}, pages 1126--1135. {PMLR},
  2017.

\bibitem{DBLP:conf/nips/SnellSZ17}
Jake Snell, Kevin Swersky, and Richard~S. Zemel.
\newblock Prototypical networks for few-shot learning.
\newblock In {\em Annual Conference on Neural Information Processing Systems
  2017, December 4-9, 2017, Long Beach, CA, {USA}}, pages 4077--4087, 2017.

\bibitem{DBLP:conf/cvpr/SunLCS19}
Qianru Sun, Yaoyao Liu, Tat{-}Seng Chua, and Bernt Schiele.
\newblock Meta-transfer learning for few-shot learning.
\newblock In {\em {IEEE} Conference on Computer Vision and Pattern Recognition,
  {CVPR} 2019, Long Beach, CA, USA, June 16-20, 2019}, pages 403--412. Computer
  Vision Foundation / {IEEE}, 2019.

\bibitem{DBLP:conf/cikm/0002CZTZG19}
Fan Zhou, Chengtai Cao, Kunpeng Zhang, Goce Trajcevski, Ting Zhong, and
  Ji~Geng.
\newblock Meta-gnn: On few-shot node classification in graph meta-learning.
\newblock In {\em Proceedings of the 28th {ACM} International Conference on
  Information and Knowledge Management, {CIKM} 2019, Beijing, China, November
  3-7, 2019}, pages 2357--2360, 2019.

\bibitem{liu2021relative}
Zemin Liu, Yuan Fang, Chenghao Liu, and Steven~CH Hoi.
\newblock Relative and absolute location embedding for few-shot node
  classification on graph.
\newblock In {\em Proceedings of the AAAI Conference on Artificial
  Intelligence}, volume~35, pages 4267--4275, 2021.

\bibitem{DBLP:conf/aaai/YaoZWJWHCL20}
Huaxiu Yao, Chuxu Zhang, Ying Wei, Meng Jiang, Suhang Wang, Junzhou Huang,
  Nitesh~V. Chawla, and Zhenhui Li.
\newblock Graph few-shot learning via knowledge transfer.
\newblock In {\em The Thirty-Fourth {AAAI} Conference on Artificial
  Intelligence, {AAAI} 2020, New York, NY, USA, February 7-12, 2020}, pages
  6656--6663, 2020.

\bibitem{DBLP:conf/cikm/DingWLSLL20}
Kaize Ding, Jianling Wang, Jundong Li, Kai Shu, Chenghao Liu, and Huan Liu.
\newblock Graph prototypical networks for few-shot learning on attributed
  networks.
\newblock In {\em The 29th {ACM} International Conference on Information and
  Knowledge Management, Virtual Event, Ireland, October 19-23, 2020}, pages
  295--304. {ACM}, 2020.

\bibitem{DBLP:conf/ijcai/YuYZ18}
Bing Yu, Haoteng Yin, and Zhanxing Zhu.
\newblock Spatio-temporal graph convolutional networks: {A} deep learning
  framework for traffic forecasting.
\newblock In {\em Proceedings of the Twenty-Seventh International Joint
  Conference on Artificial Intelligence, {IJCAI} 2018, July 13-19, 2018,
  Stockholm, Sweden}, pages 3634--3640, 2018.

\bibitem{DBLP:journals/corr/ChungGCB14}
Junyoung Chung, {\c{C}}aglar G{\"{u}}l{\c{c}}ehre, KyungHyun Cho, and Yoshua
  Bengio.
\newblock Empirical evaluation of gated recurrent neural networks on sequence
  modeling, 2014.

\bibitem{DBLP:conf/cvpr/LeaFVRH17}
Colin Lea, Michael~D. Flynn, Ren{\'{e}} Vidal, Austin Reiter, and Gregory~D.
  Hager.
\newblock Temporal convolutional networks for action segmentation and
  detection.
\newblock In {\em 2017 {IEEE} Conference on Computer Vision and Pattern
  Recognition, {CVPR} 2017, Honolulu, HI, USA, July 21-26, 2017}, pages
  1003--1012, 2017.

\bibitem{DBLP:conf/iclr/VelickovicCCRLB18}
Petar Velickovic, Guillem Cucurull, Arantxa Casanova, Adriana Romero, Pietro
  Li{\`{o}}, and Yoshua Bengio.
\newblock Graph attention networks.
\newblock In {\em 6th International Conference on Learning Representations,
  {ICLR} 2018, Vancouver, BC, Canada, April 30 - May 3, 2018, Conference Track
  Proceedings}, 2018.

\bibitem{DBLP:conf/nips/DefferrardBV16}
Micha{\"{e}}l Defferrard, Xavier Bresson, and Pierre Vandergheynst.
\newblock Convolutional neural networks on graphs with fast localized spectral
  filtering.
\newblock In {\em Annual Conference on Neural Information Processing Systems
  2016, December 5-10, 2016, Barcelona, Spain}, pages 3837--3845, 2016.

\bibitem{DBLP:conf/iclr/XuSCQC19}
Bingbing Xu, Huawei Shen, Qi~Cao, Yunqi Qiu, and Xueqi Cheng.
\newblock Graph wavelet neural network.
\newblock In {\em 7th International Conference on Learning Representations,
  {ICLR} 2019, New Orleans, LA, USA, May 6-9, 2019}, 2019.

\bibitem{DBLP:conf/iclr/LiYS018}
Yaguang Li, Rose Yu, Cyrus Shahabi, and Yan Liu.
\newblock Diffusion convolutional recurrent neural network: Data-driven traffic
  forecasting.
\newblock In {\em 6th International Conference on Learning Representations,
  {ICLR} 2018, Vancouver, BC, Canada, April 30 - May 3, 2018, Conference Track
  Proceedings}, 2018.

\bibitem{Didi_dataset}
Didi-Chuxing.
\newblock Didi chuxing gaia initiative.
\newblock \url{https://gaia.didichuxing.com}, 2020.
\newblock Accessed: 2020-02-14.

\bibitem{Du2021ADARNN}
Yuntao Du, Jindong Wang, Wenjie Feng, Sinno Pan, Tao Qin, Renjun Xu, and
  Chongjun Wang.
\newblock Adarnn: Adaptive learning and forecasting for time series.
\newblock In {\em Proceedings of the 30th ACM International Conference on
  Information \& Knowledge Management (CIKM)}, 2021.

\end{thebibliography}

\appendix

\section{Appendix}

To support the reproducibility of the results in this paper, we have released our code and data.\footnote{The implementation code and details of our model is available at Github repo: https://github.com/RobinLu1209/ST-GFSL}. Our implementation is based on torch 1.8.1, and torch-geometric 1.7.2.
All the evaluated models are implemented on a server with two CPUs (Intel Xeon E5-2630 $\times$ 2) and four GPUs (NVIDIA GTX 2080 $\times$ 4).
Here, we detail the datasets, evaluation metrics, baselines, and the training settings of ST-GFSL.

\subsection{Datasets}
\label{appendix-dataset}

We take four public dataset of traffic flow for evaluation. We adopt the same data preprocessing procedures in literature. In all those datasets, we apply Z-Score normalization for data preprocessing. The missing values are filled by the linear interpolation. To construct the road network graph, each traffic sensor or road segment is considered as a vertex and we compute the pairwise road network distances between sensors. 
The adjacency matrix of the nodes is constructed by
road network distance with a thresholded Gaussian kernel. The detailed information of each dataset is shown in Table \ref{tab:datasets}.

% \usepackage{graphicx}
\begin{table}[h]
\centering
\caption{Statistics of the traffic datasets in experiment.}
\label{tab:datasets}
\resizebox{\linewidth}{!}{%
\begin{tabular}{ccccc}
\hline
\textbf{Datasets}        & \textbf{METR-LA}     & \textbf{PEMS-BAY}    & \textbf{Didi-Chengdu}  & \textbf{Didi-Shenzhen} \\ \hline
\# Nodes        & 207         & 325         & 524          & 627           \\
\# Edges        & 1,722       & 2,694       & 1,120        & 4,845         \\
Interval  & 5 min       & 5 min       & 10 min       & 10 min        \\
Time span   & 34,272      & 52,116      & 17,280       & 17,280        \\
Mean  & 58.274      & 61.776      & 29.023       & 31.001        \\
Std & 13.128      & 9.285       & 9.662        & 10.969        \\ \hline
\end{tabular}%
}
\end{table}

\begin{itemize}[left=1em]
    \item \textbf{METR-LA}~\cite{DBLP:conf/iclr/LiYS018}: Traffic data are collected from observation sensors in the highway of Los Angeles County. We use 207 sensors and 4 months of data dated from 1st Mar 2012 until 30th Jun 2012 in the experiment. The readings of the sensors are aggregated into 5-minutes windows.
    \item \textbf{PEMS-BAY}~\cite{DBLP:conf/iclr/LiYS018}: PEMS-BAY dataset contains 6 months of data recorded by 325 traffic sensors ranging from January 1st, 2017 to June 30th, 2017 in the Bay Area. 
    \item \textbf{Didi-Chengdu}~\cite{Didi_dataset}: Traffic index dataset of Chengdu, China, provided by Didi Chuxing GAIA Initiative. We select data from January to April 2018, with 524 roads in the core urban area of Chengdu. The data collection interval is 10 minutes.
    \item \textbf{Didi-Shenzhen}~\cite{Didi_dataset}: Traffic index dataset of Shenzhen, China, provided by Didi Chuxing GAIA Initiative. We select data from January to April 2018, and it contains 627 roads in downtown Shenzhen. The data collection interval is 10 minutes.
\end{itemize}

\subsection{Evaluation Metrics}

To verify the effectiveness of the proposed algorithm, we evaluate the results of multi-step prediction. 
% For each of the four data sets, we use historical data of 12 time steps to predict the future results of 6 time steps. Due to the different time resolutions of different datasets, Didi-Shenzhen and Didi-Chengdu datasets will predict the results in the next 60 minutes, while METR-LA and PEMS-BAY datasets will predict the results in the next 30 minutes.
Two widely used metrics are applied between the multi-step prediction $\hat{y}$ and the ground truth $y$ for evaluation: Mean Absolute Error (MAE), and Root Mean Squared Error (RMSE).
\begin{itemize}[left=1em]
    \item Mean Absolute Error (MAE)
        \begin{equation*}
            MAE(y,\hat{y}) = \frac{1}{N}\sum\nolimits_{i=1}^{N}\left|\hat{y_{i}}-y_{i} \right|\text{.}
        \end{equation*}
    \item Root Mean Squared Error (RMSE)
        \begin{equation*}
            RMSE(y,\hat{y}) = \sqrt{\frac{1}{N}\sum\nolimits_{i=1}^{N}(\hat{y_{i}}-y_{i})^{2}}\text{.}
        \end{equation*}
\end{itemize}

\subsection{Baselines}
The details of the baselines are as follows:
\begin{itemize}[left=1em]
    \item \textbf{HA}: Historical Average, which formulates the traffic flow as a seasonal process, and uses average of previous seasons as the prediction. We use the few-shot sample of target city to calculate the average traffic speed of each node in one day, and use this as the baseline to predict the future values.
    \item \textbf{ARIMA}: Auto-regressive integrated moving average is a well-known model that can understand and predict future values in a time series. The model are implemented using the statsmodel python package. The orders are (3,0,1).
    \item \textbf{Target-only}: We directly use the few-shot training samples of target domain to train the model, without utilizing the training data of other source cities for knowledge transfer.
    \item \textbf{Fine-tuned (Vanilla)}: We first train the model on source datasets, and then fine-tune the model on few-shot data in target domain. Here we used all the multiple source cities for training.
    \item \textbf{Fine-tuned (ST-Meta)}: Compared with ``Fine-tuned (Vanilla)'' method, we combine the proposed parameter generation based on meta knowledge to generate non-shared parameters for the model. 
    \item \textbf{AdaRNN~\cite{Du2021ADARNN}}: A state-of-the-art transfer learning framework for non-stationary time series. This paper aims to reduce the distribution mismatch in the time series to learn an adaptive RNN-based model. In the experiment, we predict each node independently, and the traffic speed of a single node can be analyzed as a typical time series as the setting in AdaRNN. 
    \item \textbf{MAML~\cite{DBLP:conf/icml/FinnAL17}}: Model-Agnostic Meta 
    Learning (MAML), a superior meta-learning method that trains a model’s parameters such that a small number of gradient updates will lead to fast learning on a new task. It learns a better initialization model from multiple tasks to supervise the target task. 
\end{itemize}

\subsection{Training settings}
In the experiment, we use $T=12$ historical time steps to predict the traffic speed of the next $M=6$ time steps. Due to different time resolutions, Didi-Shenzhen and Didi-Chengdu datasets will predict the results in the next 60 minutes, while METR-LA and PEMS-BAY datasets will predict the next 30 minutes’.
The ST-GFSL framework is trained by Adam optimizer with learning rate decay in both inner loop and outer loop. In ST-Meta Learner, the number of GAT is set to 2, and the the GRU layer is set to 1.
Totally, there are several important hyperparameters in our model, and we set them as: the dimension of meta knowledge $d_{MK} = 16$, task learning rate $\alpha = 0.01$, meta-training rate $\beta = 0.001$, task batch number $\|\mathcal{T}\| = 5$, and sum scale factor of two loss function $\lambda = 1.5$.

In order to compare the performance of different feature extractors in the ST-GFSL framework, we also used three advanced models in this paper, which are described in detail as follows:
\begin{itemize}[left=1em]
    \item \textbf{TCN~\cite{DBLP:conf/cvpr/LeaFVRH17}}: 1D dilated convolution network-based temporal convolution network. We used the addition of two TCN models as the feature extractor. The size of the dilated convolution kernel is set to 2 and 3 respectively.
    \item \textbf{STGCN~\cite{DBLP:conf/ijcai/YuYZ18}}: Spatial temporal graph convolution network, which combines graph convolution with 1D causal convolution. The layer of STGCN block is set to 2, and use one-layer FC as the multi-step predictor.
    \item \textbf{GWN~\cite{DBLP:conf/ijcai/WuPLJZ19}}: A convolution network structure combines graph convolution with dilated casual convolution, which introduces a self-adaptive graph to capture the hidden spatial dependency. The block of GWN is set to 4, and the layer of GWN blocks is set to 2.
\end{itemize}

\end{sloppypar}
\end{document}